\documentclass[smallextended]{svjour3}       
\def\makeheadbox{\relax}

\smartqed  
\usepackage{graphicx}

\begin{document}

\title{A Review of 40 Years in Cognitive Architecture Research  \\  Core Cognitive Abilities and Practical Applications}

\author{Iuliia Kotseruba        \and
        John K. Tsotsos 
}


\institute{Iuliia Kotseruba \at
               \email{yulia\_k@cse.yorku.ca}           
           \and
          John K. Tsotsos \at
              \email{tsotsos@cse.yorku.ca}
}

\date{}

\maketitle

\begin{abstract}
In this paper we present a broad overview of the last 40 years of research on cognitive architectures. Although the number of existing architectures is nearing several hundred, most of the existing surveys do not reflect this growth and focus on a handful of well-established architectures. Thus, in this survey we wanted to shift the focus towards a more inclusive and high-level overview of the research on cognitive architectures. Our final set of 84 architectures includes 49 that are still actively developed, and borrow from a diverse set of disciplines, spanning areas from psychoanalysis to neuroscience. To keep the length of this paper within reasonable limits we discuss only the core cognitive abilities, such as perception, attention mechanisms, action selection, memory, learning and reasoning. In order to assess the breadth of practical applications of cognitive architectures we gathered information on over 900 practical projects implemented using the cognitive architectures in our list. 

We use various visualization techniques to highlight overall trends in the development of the field. In addition to summarizing the current state-of-the-art in the cognitive architecture research, this survey describes a variety of methods and ideas that have been tried and their relative success in modeling human cognitive abilities, as well as which aspects of cognitive behavior need more research with respect to their mechanistic counterparts and thus can further inform how cognitive science might progress.


\end{abstract}

\section{Introduction}
\label{intro}
The goal of this paper is to provide a broad overview of the last 40 years of research in cognitive architectures with an emphasis on the core capabilities of perception, attention mechanisms, action selection, learning, memory, reasoning and their practical applications. Although the field of cognitive architectures has been steadily expanding, most of the surveys published in the past 10 years do not reflect this growth and feature essentially the same set of a dozen most established architectures. The latest large-scale study was conducted by Samsonovich in 2010 \cite{Samsonovich2010} in an attempt to catalog the implemented cognitive architectures. His survey contains descriptions of 26 cognitive architectures submitted by their respective authors. The same information is also presented online in a Comparative Table of Cognitive Architectures\footnote{\url{http://bicasociety.org/cogarch/architectures.htm}}. In addition to surveys, there are multiple on-line sources listing cognitive architectures, but they rarely go beyond a short description and a link to the project site or a software repository.

\begin{figure*}[htpb]
  \includegraphics[width=1.00\textwidth]{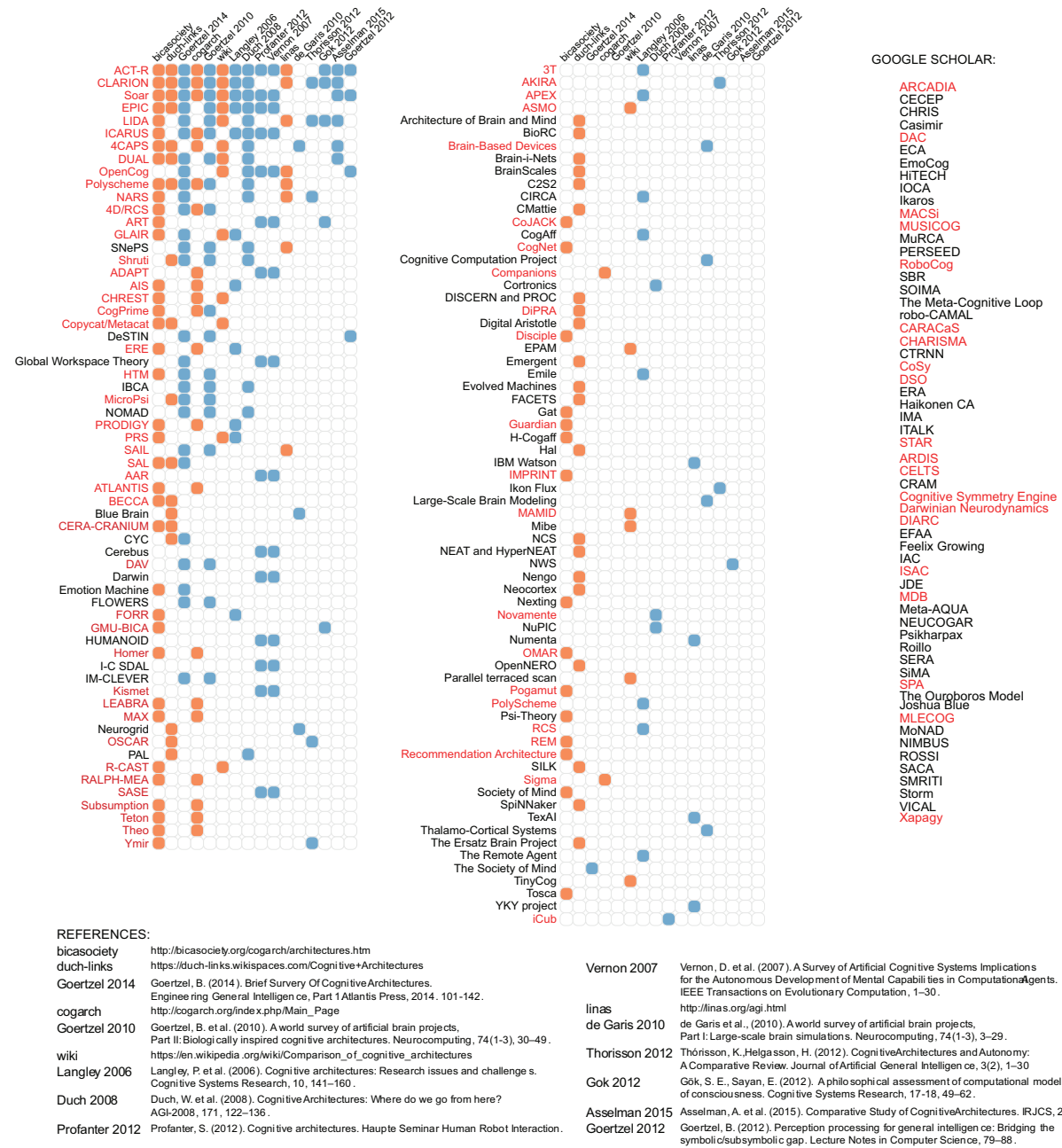}
\caption[LoF entry]{A diagram showing cognitive architectures found in literature surveys and on-line sources shown with blue and orange colors respectively. The architectures in the diagram are sorted by the total number of references in the surveys and on-line sources for each architecture. Titles of the architectures covered in this survey are shown in red. 

All visualizations in this paper are made using D3.js library (\url{https://d3js.org}) and interactive versions of the figures are available on the project website. The color palettes are generated using ColorBrewer (\url{http://colorbrewer2.org}).}
\label{fig_1_cog_arch_surveys}       
\end{figure*}

Since there is no exhaustive list of cognitive architectures, their exact number is unknown, but it is estimated to be around three hundred, out of which at least one-third of the projects are currently active. To form the initial list for our study we combined the architectures mentioned in surveys (published within the last 10 years) and several large on-line catalogs. We also included more recent projects not yet mentioned in the survey literature. Figure \ref{fig_1_cog_arch_surveys} shows a visualization of 195 cognitive architectures featured in 17 sources (surveys, on-line catalogs and Google Scholar). It is apparent from this diagram that a small group of architectures such as ACT-R, Soar, CLARION, ICARUS, EPIC, LIDA and a few others are present in most sources, while all other projects are only briefly mentioned in on-line catalogs. While the theoretical and practical contributions of the major architectures are undeniable, they represent only a part of the research in the field. Thus, in this review the focus is shifted away from the deep study of the major architectures or discussing what could be the best approach to modeling cognition, which has been done elsewhere. For example, in a recent paper by Laird et al. \cite{Laird2017} ACT-R, Soar and Sigma are compared based on their structural organization and approaches to modelling core cognitive abilities. Further, a new Standard Model of the Mind is proposed as a reference model born out of consensus between the three architectures. Our goal, on the other hand, is to present a broad, inclusive, judgment-neutral snapshot of the past 40 years of development with the goal of informing the future cognitive architecture research by presenting the diversity of ideas that have been tried and their relative success. 

\begin{figure*}[htpb]
  \includegraphics[width=1.00\textwidth]{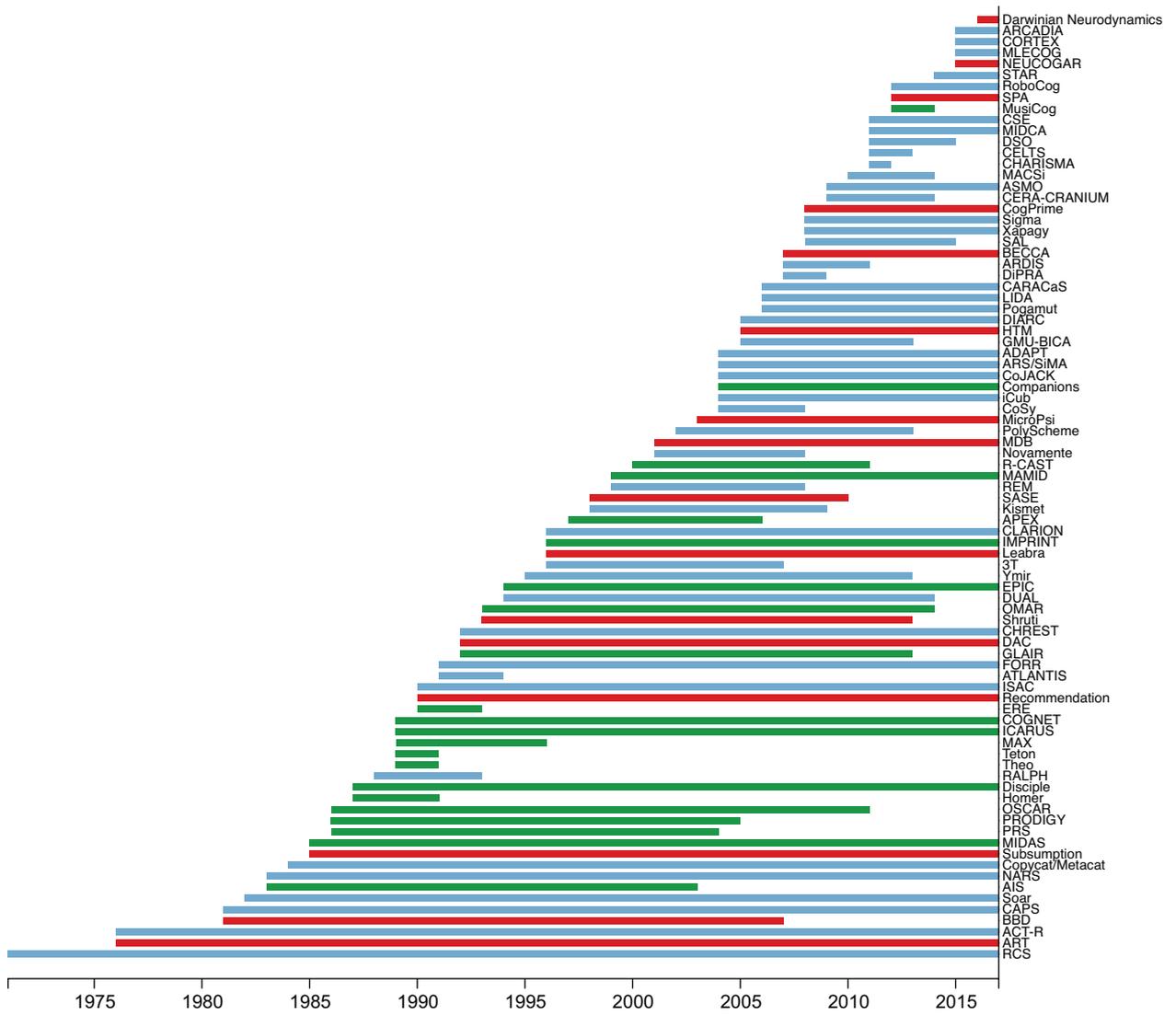}
\caption{A timeline of 84 cognitive architectures featured in this survey. Each line corresponds to a single architecture. The architectures are sorted by the starting date, so that the earliest architectures are plotted at the bottom of the figure. Since the explicit beginning and ending dates are known only for a few projects, we recovered the timeline based on the dates of the publications and activity on the project web page or on-line repository. Colors correspond to different types of architectures: symbolic (green), emergent (red) and hybrid (blue).
According to this data there was a particular interest in symbolic architectures since mid-1980s until early 1990s, however after 2000s most of the newly developed architectures are hybrid. Emergent architectures, many of which are biologically-inspired, are distributed fairly evenly in the timeline, but remain a relatively small group.}
\label{fig_2_cog_arch_timeline}       
\end{figure*}

To make this survey manageable we reduced the original list of architectures to 84 items by considering only implemented architectures with at least one practical application and several peer-reviewed publications. Even though we do not explicitly include some of the philosophical architectures such as CogAff \cite{Sloman2003}, Society of Mind \cite{Minsky1986}, Global Workspace Theory (GWT) \cite{Baars2005} and Pandemonium theory \cite{Selfridge1958}, we examine cognitive architectures heavily influenced by these theories (e.g. LIDA, ARCADIA, CERA-CRANIUM, ASMO, COGNET and Copycat/Metacat, also see discussion in Section \ref{section_5_attention}). We also exclude large-scale brain modeling projects, which are low-level and do not easily map onto the breadth of cognitive capabilities modeled by other types of cognitive architectures. Further, many of the brain models do not have practical applications, and thus do not fit the parameters of the present survey. Figure \ref{fig_2_cog_arch_timeline} shows all architectures featured in this survey with their approximate timelines recovered from the publications. Of these projects 49 are currently active\footnote{Since the exact dates for the start (and end) of the development are not specified for the majority of architectures, we use instead the first and the latest publication. If an architecture has an associated website or an online repository, we consider the project currently active, given that there was any activity (site update/code commit) within the last year. Similarly, we consider the project under the active development if there was at least one publication within the last year (2016). }.

As we mentioned earlier, the first step towards creating an inclusive and organized catalog of implemented cognitive architectures was made by Samsonovich \cite{Samsonovich2010}. His work contained extended descriptions of 26 projects with the following information: short overview, a schematic diagram of major elements, common components and features (memory types, attention, consciousness, etc.), learning and cognitive development, cognitive modeling and applications, scalability and limitations. A survey of this kind brings together researchers from several disjoint communities and helps to establish a mapping between the different approaches and terminology they use. However, the descriptive or tabular format does not allow easy comparisons between architectures. Since our sample of architectures is large, we experimented with alternative visualization strategies. 

In the following sections, we will provide an overview of the definitions of cognition and approaches to categorizing cognitive architectures. As one of our contributions, we map cognitive architectures according to their perception modality, implemented mechanisms of attention, memory organization, types of learning, action selection and practical applications. 

In the process of preparing this paper, we examined the literature widely and this activity led to a 2500 item bibliography of relevant publications. We provide this bibliography, with short summary descriptions for each paper as a supplementary material. In addition, interactive versions of all diagrams in this paper are available on-line\footnote{\url{http://jtl.lassonde.yorku.ca/project/cognitive_architectures_survey/}} and allow one to explore the data and view relevant references.

\section{What are Cognitive Architectures?}
\label{section_what_are_CA}
Cognitive architectures are a part of research in general AI, which began in the 1950s with the goal of creating programs that could reason about problems across different domains, develop insights, adapt to new situations and reflect on themselves. Similarly, the ultimate goal of research in cognitive architectures is to model the human mind, which will bring us closer to building human-level artificial intelligence. Cognitive architectures attempt to provide evidence that particular mechanisms succeed in producing intelligent behavior and thus contribute to cognitive science. Moreover, the body of work represented by the cognitive architectures, and this review, documents what methods or strategies have been tried previously (and what have not), how they have been used, and what level of success has been achieved or lessons learned, all important elements that help guide future research efforts. For AI and engineering, documentation of past mechanistic work has obvious import. But this is just as important for cognitive science, since most experimental work eventually attempts to connect to explanations of how observed human behavior may be generated and the body of cognitive architectures provides a very rich source of viable ideas and mechanisms.

According to Russel and Norvig \cite{Russell1995} artificial intelligence may be realized in four different ways: systems that think like humans, systems that think rationally, systems that act like humans, and systems that act rationally. The existing cognitive architectures have explored all four possibilities. For instance, human-like thought is pursued by the architectures stemming from cognitive modeling. In this case, the errors made by an intelligent system should match the errors typically made by people in similar situations. This is in contrast to rationally thinking systems which are required to produce consistent and correct conclusions for arbitrary tasks. A similar distinction is made for machines that act like humans or act rationally. Machines in either of these groups are not expected to think like humans, only their actions or behavior is taken into account. 

However, with no clear definition and general theory of cognition, each architecture was based on a different set of premises and assumptions, making comparison and evaluation difficult. Several papers were published to resolve the uncertainties, the most prominent being Sun's desiderata for cognitive architectures \cite{Sun2004b} and Newell's functional criteria (first published in \cite{Newell1980} and \cite{Newell1992}, and later restated by Anderson and Lebiere in \cite{Anderson2003a}). Newell's criteria include flexible behavior, real-time operation, rationality, large knowledge base, learning, development, linguistic abilities, self-awareness and brain realization. Sun's desiderata are broader and include ecological, cognitive and bio-evolutionary realism, adaptation, modularity, routineness and synergistic interaction. Besides defining these criteria and applying them to a range of cognitive architectures, Sun also pointed out the lack of clearly defined cognitive assumptions and methodological approaches, which hinder progress in studying intelligence. He also noted an uncertainty regarding essential dichotomies (implicit/explicit, procedural/declarative, etc.), modularity of cognition and structure of memory. However, a quick look at the existing cognitive architectures reveals persisting disagreements in terms of their research goals, structure, operation and application. 

Instead of looking for a particular definition of intelligence \cite{Legg2007}, it may be more practical to define it as a set of competencies and behaviors demonstrated by the system. While no comprehensive list of capabilities required for intelligence exists, several broad areas have been identified that may serve as guidance for ongoing work in the cognitive architecture domain. For example, Adams et al.\cite{Adams2012a} suggest areas such as perception, memory, attention, actuation, social interaction, planning, motivation, emotion, etc. These are further split into subareas. Arguably, some of these categories may seem more important than the others and historically attracted more attention (further discussed in Section \ref{section_10_2_application_categories}).

Implementing even a reduced set of abilities in a single architecture is a substantial undertaking. Unsurprisingly, the goal of achieving Artificial General Intelligence (AGI) is explicitly pursued by a small number of architectures, among which are Soar, ACT-R, NARS \cite{Wang2013a},  LIDA \cite{Faghihi2012a}, and several recent projects, such as SiMA (formerly ARS) \cite{Schaat2014a}, Sigma \cite{Pynadath2014} and CogPrime \cite{Goertzel2014c}. Others focus on a particular aspect of cognition, e.g. attention (ARCADIA \cite{Bridewell2015},  STAR \cite{Tsotsos2017}), emotion (CELTS \cite{Faghihi2011c}), perception of symmetry (Cognitive Symmetry Engine \cite{Henderson2013}) or problem solving (FORR \cite{Epstein2004a}, PRODIGY \cite{Epstein2004a}). There are also narrowly specialized architectures designed for particular applications, such as ARDIS \cite{Martin2009} for visual inspection of surfaces or MusiCog \cite{Maxwell2014a} for music comprehension and generation.

Different opinions can be found in the literature on what system can be considered a cognitive architecture. Laird in \cite{Laird2012c} discusses how cognitive architectures differ from other intelligent software systems. While all of them have memory storage, control components, data representation, and input/output devices, the latter provide only a fixed model for general computation. Cognitive architectures, on the other hand, must change through development and efficiently use knowledge to perform new tasks. Furthermore, he suggests toolkits and frameworks for building intelligent agents (e.g. GOMS, BDI, etc.) cannot themselves be considered cognitive architectures. (However, the authors of Pogamut, a framework for building intelligent agents, consider it a cognitive architecture as it is included in \cite{Samsonovich2010}.) Another opinion on the matter is by Sun \cite{Sun2007}, who contrasts the engineering approach taken in the field of artificial intelligence with the scientific approach of cognitive architectures. According to Sun, psychologically based cognitive architectures should facilitate the study of human mind by modeling not only the human behavior but also the underlying cognitive processes. Such models, unlike software engineering oriented "cognitive" architectures, are explicit representations of the general human cognitive mechanisms, which are essential for understanding of the mind.

In practice the term "cognitive architecture" is not as restrictive, as made evident by the representative surveys of the field. Most of the surveys define cognitive architectures as a blueprint for intelligence, or more specifically, a proposal about the mental representations and computational procedures that operate on these representations enabling a range of intelligent behaviors \cite{Duch2008,Langley2009a,Thagard2012,Profanter2012,Butt2013}. There is generally no need to justify the inclusion of the established cognitive architectures such as Soar, ACT-R, EPIC, LIDA, CLARION, ICARUS and a few others. The same applies to many biomimetic and neuroscience-inspired cognitive architectures that model cognitive processes on a neuronal level (e.g. CAPS, BBD, BECCA, DAC, SPA). It can be argued that some engineering oriented architectures pursue a similar goal as they have a set of structural components for perception, reasoning and action, and model interactions between them, which is in the spirit of the Newell's call for "unified theories of cognition" \cite{Hayes-Roth1995a}. However, when it comes to less common or new projects, the reasons for considering them are less clear. As an example, AKIRA, a framework that explicitly does not self-identify as a cognitive architecture \cite{Pezzulo2005a}, is featured in some surveys anyway \cite{Thorisson2012}. Similarly, a knowledge base Cyc \cite{Foxvog2010}, which does not make any claims about general intelligence, is presented as an AGI architecture in \cite{Goertzel2014a}.

Recently, claims have been made that deep learning is capable of “solving AI” by Google (DeepMind\footnote{\url{https://deepmind.com/}}). Likewise, Facebook AI Research (FAIR\footnote{\url{https://research.facebook.com/ai/}}) and other companies are actively working in the same direction. However, the question is where does this work stand with respect to cognitive architectures? Overall, the DeepMind research addresses a number of important issues in AI, such as natural language understanding, perceptual processing, general learning, and strategies for evaluating artificial intelligence. Although particular models already demonstrate cognitive abilities in limited domains, at this point they do not represent a unified model of intelligence.

Differently from DeepMind, Mikolov et al. from a Facebook research team explicitly discuss their work in a broader context of developing intelligent machines \cite{Mikolov2015}. Their main argument is that AI is too complex to be built all at once and instead its general characteristics should be defined first. Two such characteristics of intelligence are defined, namely, communication and learning, and a concrete roadmap are proposed for developing them incrementally. 

Currently, there are no publications about developing such a system, but overall the research topics pursued by FAIR align with their proposal for AI and also the business interests of the company. Common topics include visual processing, especially segmentation and object detection, data mining, natural language processing, human-computer interaction and network security. Since the current deep learning techniques are mainly applied to solving practical problems and do not represent a unified framework we do not include them in this review. However, given their prevalence in other areas of AI, deep learning methods will likely play some role in the cognitive architectures of the future.

To ensure both inclusiveness and consistency, cognitive architectures in this survey are selected based on the following criteria: self-evaluation as cognitive, robotic or agent architecture, existing implementation (not necessarily open-source), and mechanisms for perception, attention, action selection, memory and learning.  Furthermore, we considered the architectures with at least several peer-reviewed papers and practical applications beyond simple illustrative examples. For the most recent architectures still under development, some of these conditions were relaxed. 

An important point to keep in mind while reading this survey is that cognitive architectures should be distinguished from the models or agents that implement them. For instance, ACT-R, Soar, HTM and many other architectures are used as the basis for multiple software agents that demonstrate only a subset of capabilities declared in theory. On the other hand, some agents may implement extra features that are not available in the cognitive architecture. A good example is the perceptual system of Rosie \cite{Kirk2014}, one of the agents implemented in Soar, whereas Soar itself does not include a real perceptual system for physical sensors as part of the architecture. Unfortunately, in many cases the level of detail presented in the publications does not allow one to judge whether the particular capability is enabled by the architectural mechanisms (and is common among all models) or is custom-made for a concrete application. Therefore, to avoid confusion, we do not make this distinction and list all capabilities demonstrated by the architecture.

Note also that some architectures went through structural and conceptual changes throughout their development, notably the long-running projects such as Soar, ACT-R, CLARION, Disciple and others. For example, Soar 9 and subsequent versions added some non-symbolic elements \cite{Laird2012} and CAPS4 better accounts for the time course of cognition and individual differences compared to its predecessors 3CAPS and CAPS\footnote{\url{http://www.ccbi.cmu.edu/4CAPS/}}. Sometimes these changes are well documented (e.g. CLARION\footnote{\url{http://www.clarioncognitivearchitecture.com/release-notes}}, Soar\footnote{\url{https://soar.eecs.umich.edu/articles/articles}}), but more often they are not. In general we use the most recent variant of the architecture for analysis and assume that the features and abilities of the previous versions are retained in the new version unless there is contradicting evidence.

\section{Taxonomies of Cognitive Architectures} \label{section_3_taxonomies}
Many papers published within the last decade address the problem of evaluation rather than categorization of cognitive architectures. As mentioned earlier, Newell's criteria \cite{Newell1980,Newell1992,Anderson2003a} and Sun's desiderata \cite{Sun2004b} belong in this category. Furthermore, surveys of cognitive architectures propose various capabilities, properties and evaluation criteria, which include recognition, decision making, perception, prediction, planning, acting, communication, learning, goal setting, adaptability, generality, autonomy, problem solving, real-time operation, meta-learning, etc. \cite{Vernon2007,Langley2009a,Thorisson2012,Asselman2015}.

While these criteria could be used for classification, many of them are too fine-grained to be applied to a generic architecture. A more general grouping of architectures is based on the type of representation and information processing they implement. Three major paradigms are currently recognized: symbolic (also referred to as cognitivist), emergent (connectionist) and hybrid. Which of these representations, if any, correctly reflects the human cognitive processes, remains an open question and has been debated for the last thirty years \cite{Sun1994a,Kelley2003}. 

\begin{figure*}[htpb]
  \includegraphics[width=1.00\textwidth]{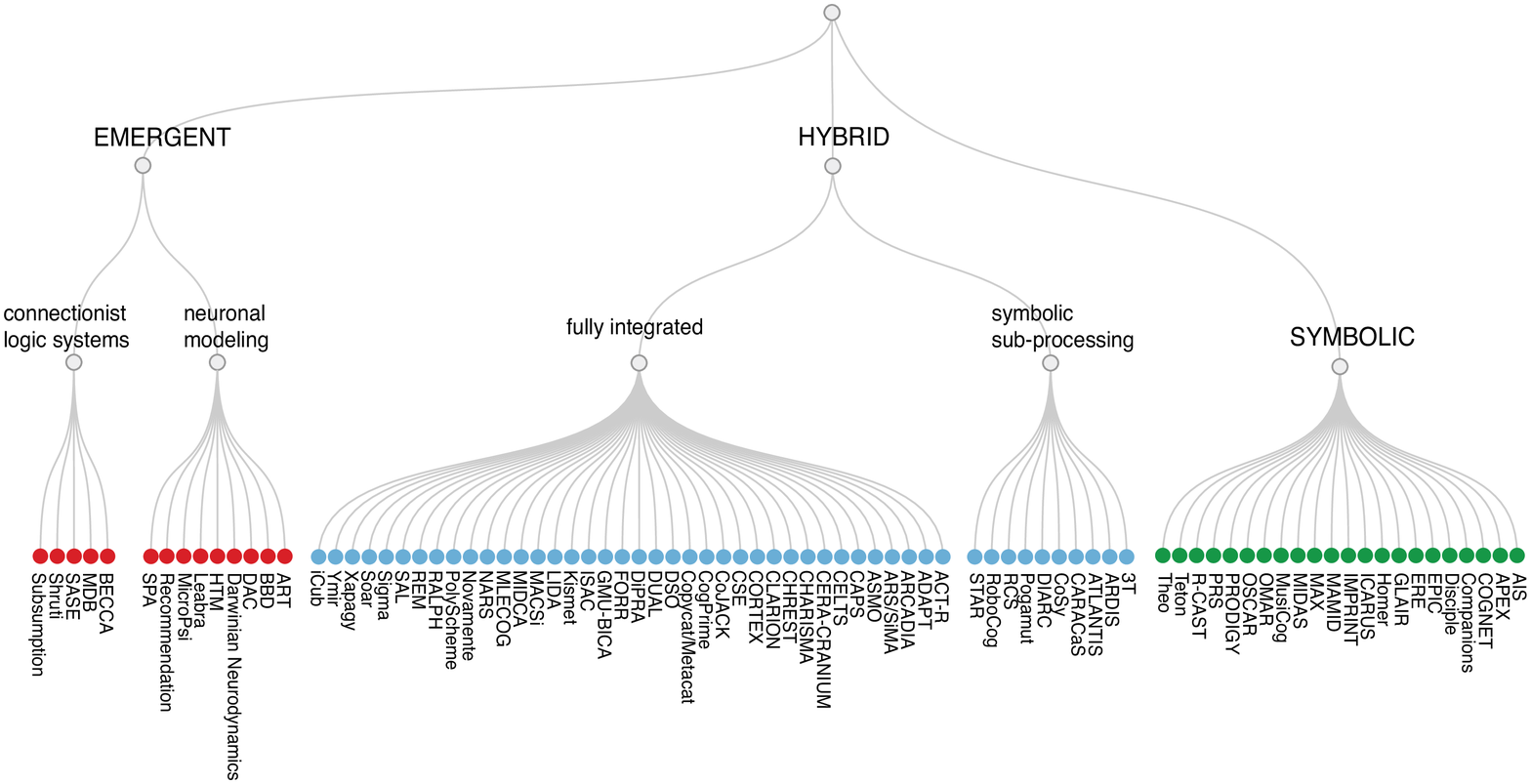}
\caption{A taxonomy of cognitive architectures based on the representation and processing. The order of the architectures within each group is alphabetical and does not correspond to the proportion of symbolic vs sub-symbolic elements (i.e. the spatial proximity of ACT-R and iCub to nodes representing symbolic and emergent architectures respectively does not imply that ACT-R is closer to symbolic paradigm and iCub is conceptually related to emergent architectures).}
\label{fig_3_cog_arch_type}       
\end{figure*}

Symbolic systems represent concepts using symbols that can be manipulated using a predefined instruction set. Such instructions can be implemented as if-then rules applied to the symbols representing the facts known about the world (e.g. ACT-R, Soar and other production rule architectures). Because it is a natural and intuitive representation of knowledge, symbolic manipulation remains very common. Although by design, symbolic systems excel at planning and reasoning, they are less able to deal with the flexibility and robustness that are required for dealing with a changing environment and for perceptual processing. 

The emergent approach resolves the adaptability and learning issues by building massively parallel models, analogous to neural networks, where information flow is represented by a propagation of signals from the input nodes. However, the resulting system also loses its transparency, since knowledge is no longer a set of symbolic entities and instead is distributed throughout the network. For these reasons, logical inference in a traditional sense becomes problematic (although not impossible) in emergent architectures.

Naturally, each paradigm has its strengths and weaknesses. For example, any symbolic architecture requires a lot of work to create an initial knowledge base, but once it is done the architecture is fully functional. On the other hand, emergent architectures are easier to design, but they must be trained in order to produce useful behavior. Furthermore, their existing knowledge may deteriorate with the subsequent learning of new behaviors. 

As neither paradigm is capable of addressing all major aspects of cognition, hybrid architectures attempt to combine elements of both symbolic and emergent approaches. Such systems are the most common in our selection of architectures (and, likely, overall). In general, there are no restrictions on how the hybridization is done and many possibilities have been explored. Multiple taxonomies of the hybridization types have been proposed \cite{Sun1996a,Hilario1997,Wermter1997,Wermter2000,Duch2007}. In addition to representation, one can consider whether the system is single or multi-module, heterogeneous or homogeneous, take into account the granularity of hybridization (coarse-grained or fine-grained), the coupling between the symbolic and sub-symbolic components, and types of memory and learning. In addition, not all the hybrid architectures explicitly address what is referred to as symbolic and sub-symbolic elements and reasons for combining them. Only a few architectures, namely ACT-R, CLARION, DUAL, CogPrime, CAPS,  SiMA, GMU-BICA and Sigma, view this integration as essential and discuss it extensively. However, we found that symbolic and sub-symbolic parts cannot be identified for all reviewed architectures due to lack of such fine-grained detail in many publications, thus we focus on representation and processing.

Figure \ref{fig_3_cog_arch_type} shows the architectures grouped according to the new taxonomy. The top level of the hierarchy is represented by the symbolic, emergent and hybrid approaches. The definitions of these terms in the literature are vague, which often leads to the inconsistent assignment of architectures to either group. There seems to be no agreement even regarding the most known architectures, Soar and ACT-R. Although both combine symbolic and sub-symbolic elements, ACT-R explicitly self-identifies as hybrid, while Soar does not \cite{Laird2012}. The surveys are inconsistent as well, for instance, both Soar and ACT-R are called cognitivist in \cite{Vernon2007} and \cite{Goertzel2010}, while \cite{Asselman2015} lists them as hybrid. 

The authors of the architectures rarely specify the types of representations used in their systems. Specifically, out of the 84 architectures we surveyed, only 34 have a self-assigned label. Fewer still provide details on the particular symbolic/subsymbolic processes or  elements. It  is universally agreed that symbols (labels, strings of characters, frames), production rules and non-probabilistic logical inference are symbolic and distributed representations such as neural networks are sub-symbolic. For example, probabilistic action selection is considered as symbolic in CARACaS \cite{Huntsberger2010}, CHREST \cite{Schiller2012} and CogPrime \cite{Goertzel2012}, but is described as sub-symbolic in ACT-R \cite{Lebiere2013}, CELTS \cite{Faghihi2011c}, CoJACK \cite{Ritter2012}, Copycat/Metacat \cite{Marshall2006} and iCub \cite{Sandini2007}. Likewise, numeric data is treated as symbolic in CAPS \cite{Just2007}, AIS \cite{Hayes-Roth1995a} and EPIC \cite{Kieras2004}, but is regarded as sub-symbolic in SASE \cite{Weng2002}. Similarly, there are disagreements regarding the status of the activations (weights or probabilities) assigned to the symbols, reinforcement learning, image data, etc.\footnote{Activations are considered symbolic only in MusiCog \cite{Maxwell2014a}, but majority of the architectures list them as sub-symbolic (e.g. CAPS \cite{Just2007}, ACT-R \cite{Moon2013,Taatgen2002}, DUAL \cite{Kokinov1994a}). Reinforcement learning is considered symbolic in CARACaS \cite{Huntsberger2010}, CHREST \cite{Schiller2012} and MusiCog \cite{Maxwell2014a}, but sub-symbolic in MAMID \cite{Reisenzein2013} and REM \cite{Murdock2001}. Image data is sub-symbolic in MAMID \cite{Reisenzein2013} and SASE \cite{Weng2002} and non-symbolic (iconic) in Soar \cite{Laird2012} and RCS \cite{Albus2005}.}

To avoid inconsistent grouping, we did not rely on self-assigned labels or conflicting classification of elements found in the literature. We assume that explicit symbols are atoms of symbolic representation and can be combined to form meaningful expressions. Such symbols are used for inference or syntactical parsing. Sub-symbolic representations are generally associated with the metaphor of a neuron. A typical example of such representation is the neural network, where knowledge is encoded as a numerical pattern distributed among many simple neuron-like objects. Weights associated with units affect processing and are acquired by learning. For our classification, we assume that anything that is not an explicit symbol and processing other than syntactic manipulation is sub-symbolic (e.g. numeric data, pixels, probabilities, spreading activations, reinforcement learning, etc.). Hybrid representation combines any number of elements from both representations.

Given these definitions, we assigned labels to all the architectures and visualized them as shown in Figure \ref{fig_3_cog_arch_type}. We distinguish between two groups in the emergent category: neuronal models, which implement models of biological neurons, and connectionist logic systems, which are closer to artificial neural networks. Within the hybrid architectures we separate symbolic sub-processing as a type of hybridization where a symbolic architecture is combined with self-contained module performing sub-symbolic computation, e.g. for processing sensory data. Other types of functional hybrids also exist, for example, co-processing, meta-processing and chain-processing, however, these are harder to identify from the limited data available in the publications and are not included. The fully integrated category combines all other types of hybrids. 

The architectures in the symbolic sub-processing group include at least one sub-symbolic module for sensory processing, while the rest of the knowledge and processing is symbolic. For example, 3T, ATLANTIS, RCS, DIARC, CARACaS and CoSy are robotic architectures, where a symbolic planning module determines the behavior of the system and one or more modules are used to process visual and audio sensory data using techniques like neural networks, optical flow calculation, etc. Similarly, ARDIS and STAR combine symbolic knowledge base and case-based symbolic reasoning with image processing algorithms.

The fully integrated architectures use a variety of approaches for combining different representations. ACT-R, Soar, CAPS, Copycat/Metacat, CHREST, CHARISMA, CELTS, CoJACK, CLARION, REM, NARS and Xapagy combine symbolic concepts and rules with sub-symbolic elements such as activation values, spreading activation, stochastic selection process, reinforcement learning, etc. On the other hand, DUAL consists of a large number of highly interconnected hybrid agents, each of which has a symbolic and sub-symbolic representation, implementing integration at a micro-level. In the case of SiMA hybridization is done on multiple levels: neural networks are used to build neurosymbols from sensors and actuators (connectionist logic systems) and the top layer is a symbol processing system (symbolic subprocessing). For the rest of the architectures, it is hard to clearly define hybridization strategy. For instance, many architectures are implemented as a set of interconnected competing and cooperating modules, where individual modules are not restricted to a particular representation (Kismet, LIDA, MACSi, ISAC, iCub, GMU-BICA, CORTEX, ASMO, CELTS, PolyScheme, FORR).  

Another hybridization strategy is to combine two different stand-alone architectures with complementary features. Even though additional effort is required to build interfaces for communication between them, this approach takes advantage of the strengths of each architecture. A good overview of conceptual and technical challenges involved in creating an interface between cognitive and robotic architectures is given in \cite{Scheutz2013a}. The proposed framework is demonstrated with two examples: ACT-R/DIARC and ICARUS/DIARC integration.  Other attempts found in the literature are CERA-CRANIUM/Pogamut \cite{Arrabales2009a} and CERA-CRANIUM/Soar \cite{LlarguesAsensio2014} hybrids for playing video games and IMPRINT/ACT-R  for human error modeling \cite{Lebiere2002}.

Few long-standing projects (based on the number of publications) also exist. A canonical example is SAL, which comprises ACT-R and Leabra architectures \cite{Jilk2008}. Here ACT-R is used to guide learning of Leabra models.  In the robotic  architecture ADAPT, Soar is utilized for control, whereas separate modules are responsible for modeling a 3D world from sensor information and for visual processing \cite{Benjamin2013}. 

Compared to hybrids, emergent architectures form a more uniform group. As mentioned before, the main difference between neuronal modeling and connectionist logic systems is in their biological plausibility. All systems in the first group implement particular neuron models and aim to accurately reproduce the low-level brain processes. The systems in the second group are based on artificial neural networks. Despite being heavily influenced by neuroscience and the ability to closely model certain elements of human cognition, the biological plausibility of these models is not claimed by their authors.

In conclusion, as can be seen in Figure \ref{fig_3_cog_arch_type}, hybrid architectures are the most numerous and diverse group, showing the tendency to grow even more, thus confirming a prediction made almost a decade ago \cite{Duch2008}. Hybrid architectures form a continuum between emergent and symbolic systems depending on the proportions and roles played by symbolic and sub-symbolic components. Although quantitative analysis of this space is not feasible, it is possible to crudely subdivide it. For instance, some architectures such as CogPrime and Sigma are conceptually closer to emergent systems as they share many properties with the neural networks. On the other hand, REM, CHREST, and RALPH, as well as the architectures implementing symbolic sub-processing, e.g. 3T and ATLANTIS, are very much within the cognitivist paradigm. These architectures are primarily symbolic but may utilize probabilistic reasoning and learning mechanisms.

\section{Perception}
\begin{figure*}[th!]
  \includegraphics[width=0.95\textwidth]{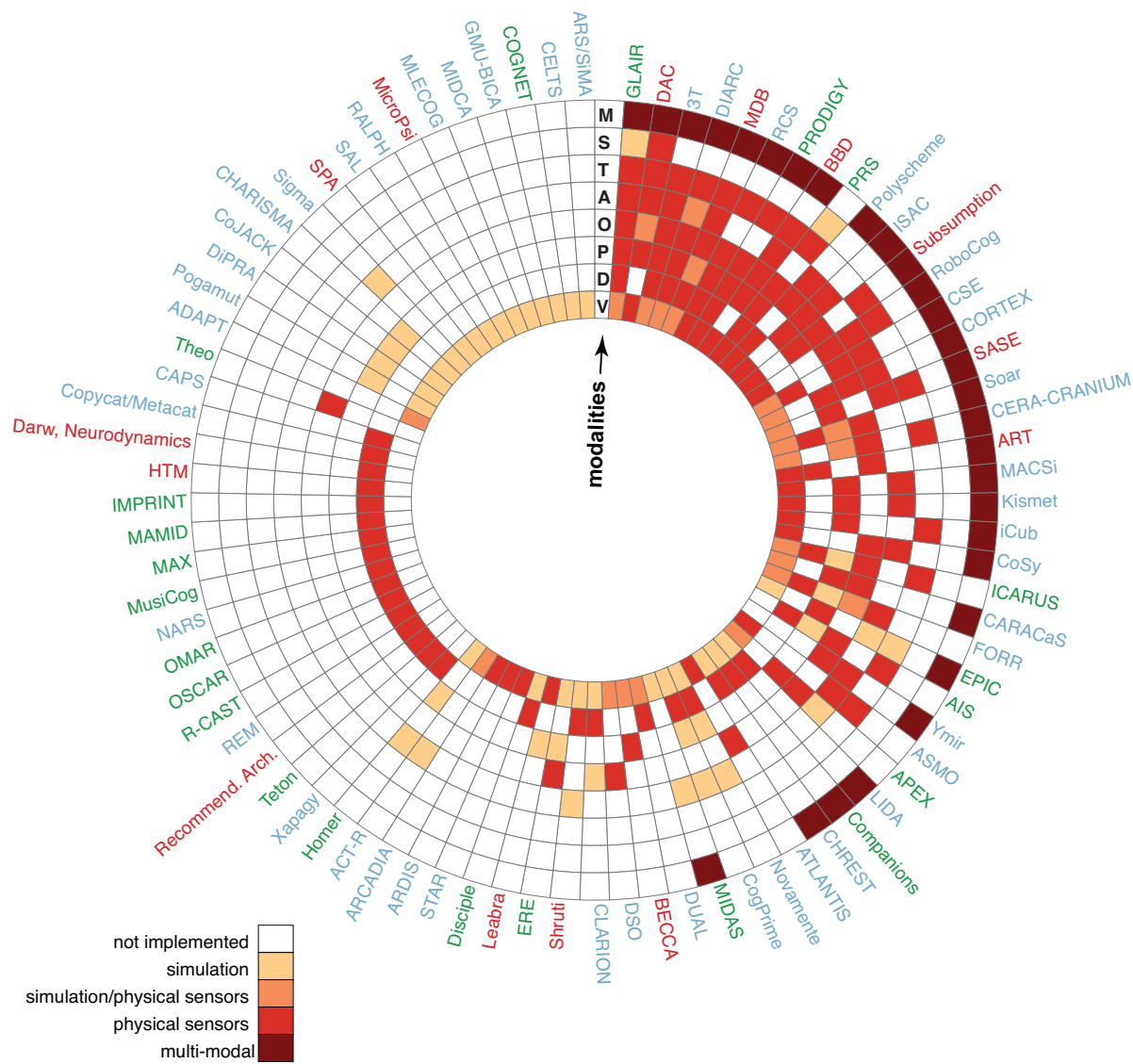}
\caption{A diagram showing sensory modalities of cognitive architectures. Radial segments correspond to cognitive architectures and each track corresponds to a modality. Modalities are ordered based on their importance (calculated as a number of architectures that implement this modality): vision (V), symbolic input (D), proprioception (P), other sensors (O), audition (A), touch (T), smell (S) and multi-modal (M). Modalities plotted closer to the center of the diagram are more common. The following coloring convention was used for the individual segments: white (the modality is not implemented), yellow (the modality is simulated), orange (sensory modality demonstrated both in real and simulated domains) and red (physical sensors used). Architecture labels are color-coded as follows: green - symbolic, blue - hybrid, red - emergent.  Architectures are ordered clockwise from 12 o'clock based on how many different sensory modalities they implement and to what extent (scores used for ranking range from 0 when the modality is not implemented to 4 for multi-modal perception).}
\label{fig_4_cog_arch_perception}       
\end{figure*}

Regardless of its design and purpose, an intelligent system cannot exist in isolation and requires an input to produce any behavior. Although historically the major cognitive architectures focused on higher-level reasoning, it is evident that perception and action play an important role in human cognition \cite{Anderson2004}.  

 Perception can be defined as a process that transforms raw input into the system's internal representation for carrying out cognitive tasks. Depending on the origin and properties of the incoming data, multiple sensory modalities are distinguished. For instance, five most common ones are vision, hearing, smell, touch and taste. Other human senses include proprioception, thermoception, nocioception, sense of time, etc. Naturally, cognitive architectures implement some of these as well as other modalities that do not have a correlate among human senses such as symbolic input (using a keyboard or graphical user interface (GUI)) and various sensors such as LiDAR, laser, IR, etc. 

Depending on its cognitive capabilities, an intelligent system can take various amounts and types of data as perceptual input. Thus, this section will investigate the diversity of the data inputs used in cognitive architectures, what information is extracted from these sources and how it is applied. The visualization in Figure \ref{fig_4_cog_arch_perception} addresses the first part of the question by mapping the cognitive architectures to the sensory modalities they implement: vision (V), hearing (A), touch (T), smell (S), proprioception (P) as well as the data input (D), other sensors (O) and multi-modal (M)\footnote{The taste modality is omitted as it is only featured in a single architecture, Brain-Based Devices (BBD), where it is simulated by measuring the conductivity of the object \cite{Krichmar2002}. }. Note that the data presented in the diagram is aggregated from many publications and most architectures do not demonstrate all listed sensory modalities in a single model.  

Several observations can be made from the visualization in Figure \ref{fig_4_cog_arch_perception}. For example, vision is the most commonly implemented sense, however, more than half of the architectures use simulation for visual input instead of the physical camera. Modalities such as touch and proprioception are mainly used in the physically embodied architectures. Some senses remain relatively unexplored, for example, sense of smell is only featured in three architectures (GLAIR \cite{Shapiro2005}, DAC \cite{Mathews2009} and PRS \cite{Taylor1996}). Overall, symbolic architectures by design have limited perceptual abilities and tend to use direct data input as the only source of information (see left side of the diagram).  On the other hand, hybrid and emergent architectures (located mainly in the right half of the diagram) implement a broader variety of sensory modalities both with simulated and physical sensors. However, regardless of its source, incoming sensory data is  usually not usable in the raw form (except, maybe, for symbolic input) and requires further processing. Below we discuss various approaches to the problem of efficient and adequate perceptual processing in cognitive architectures. 

\subsection{Vision} \label{section_4_vision}
For a long time, vision was viewed as the dominating sensory modality based on the available experimental evidence \cite{Posner1976}. While recent works suggest a more balanced view of human sensory experience \cite{Stokes2014}, cognitive architectures research remains fairly vision-centric, as it is the most studied and the most represented sensory modality. Even though in robotics various non-visual sensors (e.g. sonars, ultrasonic distance sensors) and proprioceptive sensors (e.g. gyroscopes, compasses) are used for solving visual tasks such as navigation, obstacle avoidance and search, visual input accounts for more than half of all implemented input modalities. 

According to Marr \cite{Marr2010} visual processing is composed of three different stages: early, intermediate and late. Early vision is data-driven and involves parallel processing of the visual scene and extracts simple elements, such as color, luminance, shape, motion, etc. Intermediate vision groups elements into regions, which are then further processed in the late stage to recognize objects and assign meaning to them using available knowledge. Although not mentioned by Marr, visual attention mechanisms, emotion and reward systems also influence all stages of visual processing \cite{Tsotsos2011}. Thus, perception and cognition are tightly interconnected starting throughout all levels of processing.

We base our analysis of visual processing in the cognitive architectures on image understanding stages described in \cite{Tsotsos1992}. These stages include: 1) detection and grouping of intensity-location-time values (results in edges, regions, flow vectors); 2) further grouping of edges, regions, etc. (produces surfaces, volumes, boundaries, depth information); 3) identification of objects and their motions; 4) building object-centered representations for entities; 5) assigning labels to objects based on the task; 6) inference of spatiotemporal relationships among entities\footnote{The last stage – forming consistent internal descriptions - from \cite{Tsotsos1992} is omitted. In cognitive architectures, such representations would be distributed among various reasoning and memory modules, making proper evaluation infeasible.}. Here only stage 1 represents early vision, and all subsequent stages require additional task or world knowledge. Already at stage 2, grouping of features may be facilitated by the viewpoint information and knowledge of the specific objects being viewed. Finally, the later stages involve spatial reasoning and operate on high-level representations abstracted from the results of early and intermediate processing.

Note that in computer vision research many of these image understanding stages are implemented implicitly with deep learning methods. Given the success of deep learning in many applications it is surprising to see that very few cognitive architectures employ it. Some applications of deep learning to simple vision tasks can be found in CogPrime \cite{Goertzel2013b}, LIDA \cite{Madl2015a}, SPA \cite{Stewart2013} and BECCA \cite{Rohrer2013}.

\begin{figure*}
  \includegraphics[width=1.00\textwidth]{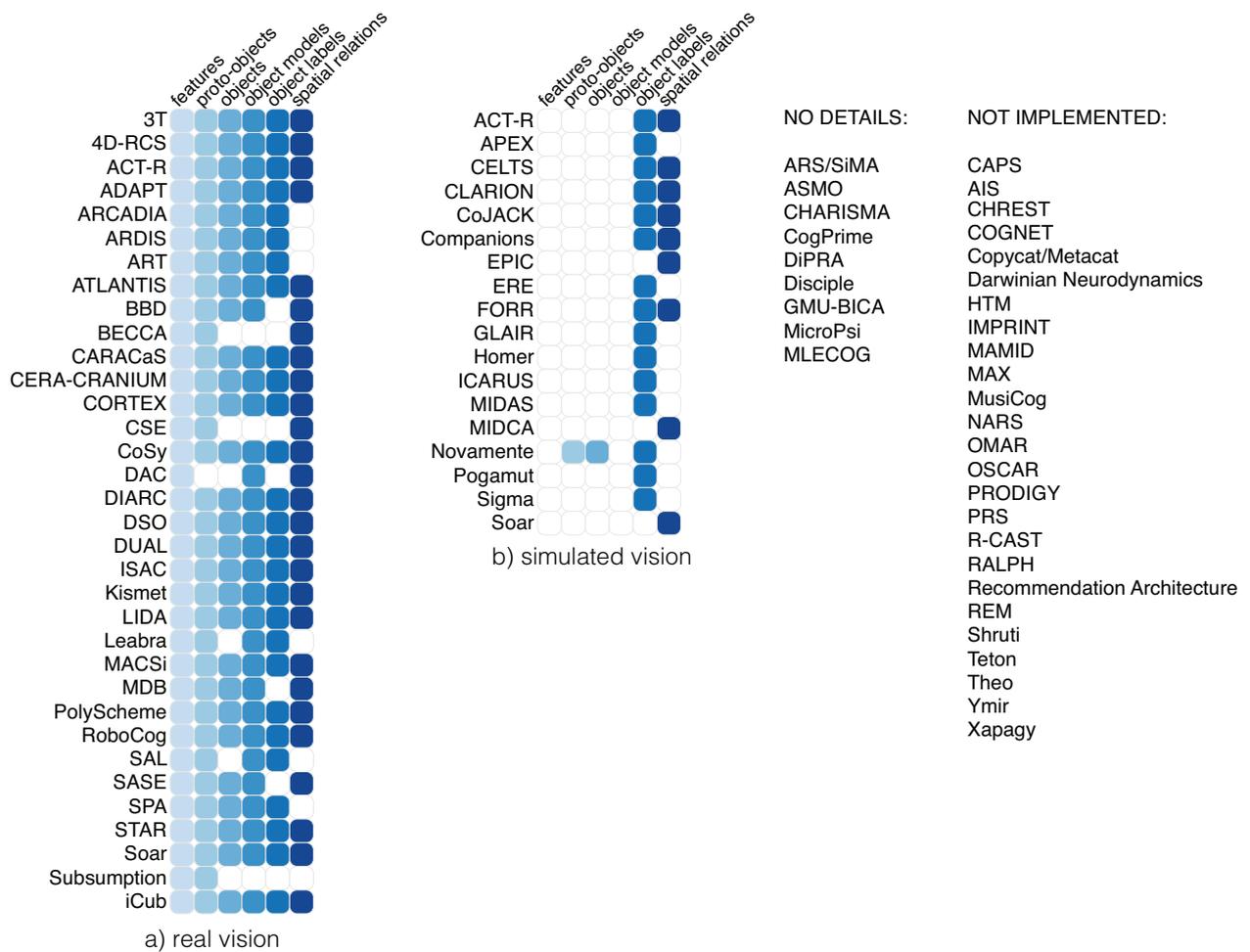}
\caption{A diagram showing the stages of real and simulated visual processing implemented by the cognitive architectures. The stages are ordered from early to late processing: 1) features, 2) proto-objects, 3) objects, 4) object models, 5) object labels, 6) spatial relations.  Different shades of blue are used to indicate processes that belong to early, intermediate and late vision. The architectures with real and simulated vision are shown in left and right columns respectively. The order within each column is alphabetical. In some architectures some type of visual process is implemented, however, the publications do not provide a sufficient amount of technical detail for our analysis ("No details" list). Architectures in the "Not implemented" list do not implement vision explicitly, although some may use other sensors (e.g. sonars and bumpers) for visual tasks such as recognition and navigation.}
\label{fig_5_cog_arch_vision}       
\end{figure*}

Diagrams in Figure \ref{fig_5_cog_arch_vision} show stages of processing implemented for real and simulated vision. We assume that the real vision systems only receive pixel-level input with no additional information (e.g. camera parameters, locations and features of objects, etc.). The images should be produced by a physical camera, but the architecture does not need to be connected to a physical sensor (i.e. if the input is a dataset of images or previously recorded video, it is still considered real vision). The simulated vision systems generally omit early and intermediate vision and receive input in the form that is suitable for the later stages of visual processing (e.g. symbolic descriptions for shape and color, object labels, coordinates, etc.).  Technically, any architecture that does not have native support for real vision or other perceptual modalities, may be extended with an interface that connects it to sensors and pre-processes raw data into a more suitable format (e.g. Soar \cite{Mohan2012} and ACT-R \cite{Trafton2011}).

The illustration in Figure \ref{fig_5_cog_arch_vision} shows what image interpretation stages are implemented, but it does not reflect the complexity and extent of such processing as there are no quantitative criteria for evaluation. In the remainder of this section we will provide brief descriptions of the visual processing in various architectures.

\subsection{Vision using physical sensors}

The majority of the architectures implementing all stages of visual processing are physically embodied and include robot control, biologically inspired and biomimetic architectures. The architectures in the first group (3T, ATLANTIS and CARACaS) operate in realistic unstructured environments and approach vision as an engineering problem. Early vision (step 1) usually involves edge detection and disparity estimation. These features are then grouped (step 2) into blobs with similar features (color, depth, etc.), which are resolved into candidate objects with centroid coordinates (step 3). Object models are learned off-line using machine learning techniques (step 4) and can be used to categorize candidate objects (step 5). For example, in RCS architecture the scene is reconstructed from stereo images and features of the traversable ground are learned from LiDAR data and color histograms \cite{Hong2002}. Similarly, the ATLANTIS robot architecture constructs a height map from stereo images and uses it to identify obstacles \cite{Miller1991}. The CARACaS architecture also computes a map for hazard avoidance from stereo images and range sensors. The locations and types of obstacles are placed on the map relative to the robot itself. Its spatial reasoning (step 6) is limited to determining the distance to the obstacles or target locations for navigation \cite{Huntsberger2011c}. 

Biologically inspired architectures also make use of computer vision algorithms and follow similar processing steps. For instance, neural networks for object detection (RCS \cite{Albus2006a}, DIARC \cite{Scheutz2004}, Kismet \cite{Breazeal2002b}), SIFT features for object recognition (DIARC \cite{Schermerhorn2006}), SURF features, AdaBoost learning and a mixture of Gaussians for hand detection and tracking (iCub \cite{Ciliberto2011}), Kinect and LBP features in conjunction with SVM classifier to find people and determine their age and gender (RoboCog and CORTEX \cite{Martinez-Gomez2014,Bandera}). 

In these architectures vision is more intertwined with memory and control systems and some steps in visual processing are explicitly related to the human visual system. One such example is saliency, which models the ability to prioritize visual stimuli based on their features or relevance to the task. As such, saliency is used to find regions of interest in the scene (Kismet \cite{Breazeal2001b}, ARCADIA \cite{Bridewell2015}, DIARC \cite{Scheutz2012}, iCub \cite{Leitner2013}, STAR \cite{Kotseruba2016}).  Ego-sphere, a structure found in some robotic architectures, mimics functions of the hippocampus in the integration of sensory information with action, although not in a biologically plausible way \cite{Peters2001}. Essentially, ego-sphere forms a virtual dome surrounding the robot, onto which salient objects and events are mapped. Various implementations of this concept are included in RCS \cite{Albus1994a}, ISAC \cite{Kawamura2008}, iCub \cite{Vernon2010a} and MACSi \cite{Oudeyer}.

The architectures in the third subgroup pursue biologically plausible vision. One of the most elaborate examples is the Leabra vision system called LVis \cite{OReilly2013} based on the anatomy of the ventral pathway of the brain. It models the primary visual cortex (V1), extrastriate areas (V2, V4) and the inferotemporal (IT) cortex. Computation in these areas roughly corresponds to early and intermediate processing steps on the diagram. LVis possesses other features of the human visual system, such as larger receptive fields of neurons in higher levels of the hierarchy, reciprocal connections between the layers and recurrent inhibitory dynamics that limit activity levels across layers \cite{Wyatte2012}. The visual systems of Darwin VIII (BBD) \cite{Seth2004a}, SPA (Spaun) \cite{Eliasmith2012} and ART \cite{Grossberg2007} are also modeled on the primate ventral visual pathway. 

The SASE architecture does not replicate the human visual system as closely. Instead, it uses a hierarchical neural network with localized connections, where each neuron gets input from a restricted region in the previous layer. The sizes of receptive fields within one layer are the same and increase at higher levels \cite{Zhang2002}. This system was tested on a SAIL robot in an indoor navigation scenario \cite{Weng2002a}. A similar approach to vision is implemented in MDB \cite{Duro2010a}, BECCA \cite{Rohrer2009} and DAC \cite{Mathews2009}.

Note that although emergent systems do not explicitly assign labels to objects, they are capable of forming representations of spatial relations between the objects in the scene and use these representations for visual tasks like navigation (BBD \cite{Fleischer2009}, BECCA \cite{Rohrer2009}, DAC \cite{Mathews2009}, MDB \cite{Duro2010a}, SASE \cite{Weng2007a}).

\subsection{Simulated vision}

As evident from the diagram in Figure \ref{fig_5_cog_arch_vision}, most of the simulations support only the late stages of visual processing. The simplest kind of simulation is a 2D grid of cells populated by objects, e.g. NASA TileWorld \cite{Systems1991} used by ERE \cite{Drummond1990} and PRS \cite{Kinny1992},  Wumpus World for GLAIR agents \cite{Shapiro2005}, 2D maze for FORR agent Ariadne \cite{Epstein1995c} and tribal simulation designed for CLARION social agents \cite{Sun2012a}. An agent in grid-like environments often has a limited view of the surroundings, restricted to few cells in each direction. Blocks world is another classic domain, where the general task is to build stacks of blocks of various shapes and colors (ACT-R \cite{Kennedy2006}, ICARUS \cite{Langley1990a}, MIDCA \cite{Cox2013a}. Despite their varying complexity and purpose, different simulations usually provide the same kinds of data about the environment: objects, their properties (color, shape, label, etc.), locations and properties of the agent itself, spatial relations between objects and environmental factors (e.g. weather and wind direction). Such simulations mainly serve as a visualization tool and are not too far removed from the direct data input since little to none sensory processing is needed. 

More advanced simulations represent the scene as polygons with color and 3D coordinates of corners, which have to be further processed to identify objects (Novamente \cite{Heljakka2007}). Otherwise, the visual realism of 3D simulations is mostly for aesthetics and sensory, as the information is available directly in symbolic form (e.g. CoJACK \cite{Ritter2009}, Pogamut \cite{Bida2012}). 

As mentioned earlier, the diagram in Figure \ref{fig_5_cog_arch_vision} does not reflect the differences in the complexity of the environments or capabilities of the individual architectures. However, there is a lot of variation in terms of size and realism among environments for the embodied cognitive architectures. For example, a planetary rover controlled by ATLANTIS performs cross-country navigation in an outdoor rocky terrain \cite{Matthies1992}, the salesman robot Gualzru (CORTEX \cite{Bandera}) moves around a large room full of people and iCub (MACsi \cite{Ivaldi2012}) recognizes and picks up various toys from a table. On the other hand, simple environments with no clutter or obstacles are also used in the cognitive architectures research (BECCA \cite{Rohrer2009}, MDB \cite{Bellas2006}). In addition, color-coding objects is a common way of simplifying visual processing. For instance, ADAPT tracks a red ball rolling on the table \cite{Benjamin2013} and DAC orients itself towards targets marked with different colors \cite{Maffei2015}. 

Furthermore, most reviewed visual systems can recognize only a handful of different object categories. In our selection, only Leabra is able to distinguish dozens of object categories \cite{Wyatte2012}. The quality of visual processing is greatly improved with the spread of available software toolkits such as OpenCV, Cloud Point Library or Kinect API. But not as much progress has been made within systems that try to model general purpose and biologically plausible visual systems. Currently, their applications are limited to controlled environments.

\subsection{Audition}

Audition is a fairly common modality in the cognitive architectures as sound or voice commands are typically used to guide an intelligent system or to communicate with it. Since the auditory modality is purely functional, many architectures resort to using available speech-to-text software rather than develop models of audition. Among the few architectures modeling auditory perception are ART, ACT-R, SPA and EPIC. For example, ARTWORD and ARTSTREAM were used to study phonemic integration \cite{Grossberg1999} and source segregation (cocktail party problem) \cite{Grossberg2004} respectively. A model of music interpretation was developed with ACT-R \cite{Chikhaoui2009}.

Using dedicated software for speech processing and communication helps to achieve a high degree of complexity and realism. For example, in robotics applications it allows a salesman robot to have scripted interaction with people in a crowded room (CORTEX \cite{Bandera}) or dialog about the objects in the scene in a subset of English (CoSy \cite{Lison2008}). A more advanced application involves using speech recognition for the task of ordering books from the public library by phone (FORR \cite{Epstein2012c}). Other systems using off-the-shelf speech processing software include the PolyScheme \cite{Trafton2005b} and ISAC \cite{Peters2001}.

In our sample of architectures, most effort is directed at natural language processing, i.e. linguistic and semantic information carried by the speech (further discussed in Section \ref{section_10_2_application_categories}), and much less attention is paid to the emotional content, (e.g. loudness, speech rate, and intonation). Some attempts in this direction are made in social robotics. For example, the social robot Kismet does not understand what is being said to it but can determine approval, prohibition or soothing based on the prosodic contours of the speech \cite{Breazeal2002c}. The Ymir architecture also has a prosody analyzer combined with a grammar-based speech recognizer that can understand a limited vocabulary of 100 words \cite{Thorisson1999a}. Even the sound itself can be used as a cue, for example, the BBD robots can orient themselves toward the source of a loud sound \cite{Seth2004a}.

\subsection{Symbolic input}

The symbolic input category in Figure \ref{fig_4_cog_arch_perception} combines several input methods which do not fall under physical sensors or simulations. These include input in the form of text commands and data, and via a GUI. Text input is typical for the architectures performing planning and logical inference tasks (e.g. NARS \cite{Wang2013a}, OSCAR \cite{Pollock1993}, MAX \cite{Kuokka1989}, Homer \cite{Vere1990}). Text commands are usually written in terms of primitive predicates used in the architecture, so no additional parsing is required.

Although many architectures have tools for visualization of results and the intermediate stages of computation, interactive GUIs are less common. They are mainly used in human performance research to simplify input of the expert knowledge and to allow multiple runs of the software with different parameters (IMPRINT \cite{Mitchell2009c}, MAMID \cite{Hudlicka2000}, OMAR \cite{Deutsch1998}, R-CAST \cite{From2011}).

The data input can be in text or any other format (e.g. binary or floating point arrays) and is primarily used for the categorization and classification applications (e.g. HTM \cite{Lavin2015}, CSE \cite{Henderson2013}, ART \cite{Carpenter1991}).

\subsection{Other modalities}

Other sensory modalities such as touch, smell and proprioception are trivially implemented with a variety of sensors (physical and simulated). For instance, many robotic platforms are equipped with touch-sensitive bumpers for an emergency stop in case of hitting an obstacle (e.g. RCS \cite{Albus2006}, AIS \cite{Hayes-Roth1993}, CERA-CRANIUM \cite{Arrabales2009b}, DIARC \cite{Langley2011a}). Similarly, proprioception, i.e. sensing of the relative positions of the body parts and effort being exerted, can be either simulated or provided by force sensors (3T \cite{Firby1995a}, iCub \cite{Pattacini2010}, MDB \cite{Bellas2010}, Subsumption \cite{Brooks1990}), joint feedback (SASE \cite{Huang2007}, RCS \cite{Bostelman2006}), accelerometers (Kismet \cite{Breazeal2003c}, 3T \cite{Wong1995}), etc.

\subsection{Multi-modal perception}

In the previous sections we have considered all sensory modalities in isolation. In reality, however, the human brain receives a constant stream of information from different senses and integrates it into a coherent world representation. This is also true for cognitive architectures, as nearly half of them have more than two different sensory modalities (see Figure \ref{fig_4_cog_arch_perception}). As mentioned in the beginning of this section, not all of these modalities may be present in a single model and most architectures utilize only two different modalities simultaneously, e.g. vision and audition, vision and symbolic input or vision and range sensors. With few exceptions, these architectures are embodied and essentially perform what is known as feature integration in cognitive science \cite{Zmigrod2013} or sensor data fusion in robotics \cite{Khaleghi2013}. Evidently, it is possible to use different sensors without explicitly combining their output, as demonstrated by the Subsumption architecture \cite{Flynn1989}, but this approach is rather uncommon. 

Multiple sensory modalities improve the robustness of perception through complementarity and redundancy but in practice using many different sensors introduces a number of challenges, such as incomplete/spurious/conflicting data, data with different properties (e.g. dimensionality or value ranges), the need for data alignment and association, etc. These practical issues are well researched by the robotics community, however, no universal solutions have been proposed. Eventually, every solution must be custom-made for a particular application, a common approach taken by the majority of cognitive architectures. There is, unfortunately, little technical information in the literature to determine the exact techniques used and connect them to the established taxonomies (e.g. from a recent survey \cite{Khaleghi2013}). 

Overall, any particular implementation of sensory integration depends on the representations used for reasoning and the task. In typical robotic architectures with symbolic reasoning, data from various sensors is processed independently and mapped onto a 3D map centered on the agent that can be used for navigation (CaRACAS \cite{Elkins2007}, CoSy \cite{Christensen2009}). In social robotics applications, as we already mentioned, the world representation can take the form of an ego-sphere around the agent that contains ego-centric coordinates and properties of the visually detected objects, which are associated to the location of the sound determined via triangulation (ISAC \cite{Peters2001a}, MACsi \cite{Anzalone2013}). In RCS, a model with a hierarchical structure, there is a sensory processing module at every level of the hierarchy with a corresponding world representation (e.g. pixel maps, 3D models, state tables, etc.) \cite{Schlenoff2005}.

Some architectures do not perform data association and alignment explicitly. Instead, sensor data and feature extraction (e.g. coordinates of objects from the camera and distances to the obstacles from laser) are done independently and concurrently. The extracted information is then directly added to the working memory, the blackboard or an equivalent structure (see Section \ref{subsection_working_memory} for more details). Any ambiguities and inconsistencies are resolved by higher-order reasoning processes. This is a common approach in distributed architectures, where independent modules concurrently work towards achieving a common goal (e.g. CERA-CRANIUM \cite{Arrabales2009b}, Polyscheme \cite{Cassimatis2004}, RoboCog \cite{Bustos2013}, Ymir \cite{Thorisson1997a} and LIDA \cite{Tobergte2015}). 

In many biologically inspired architectures the association between the readings of different sensors (and actions) is learned. For example, DAC uses Hebbian learning to establish data alignment for mapping neural representations of different sensory modalities to a common reference frame, mimicking the function of the superior colliculus of the brain \cite{Mathews2012}. ART integrates visual and ultrasonic sensory information via neural fusion (ARTMAP network) for mobile robot navigation \cite{Martens}. Likewise, MDB uses neural networks to learn the world model from sensor inputs and a genetic algorithm to adjust the parameters of the networks \cite{Bellas2010}.

All approaches mentioned so far bear some resemblance to human sensory integration as they use the spatial and temporal proximity and/or learning to combine and disambiguate multi-modal data. But overall, only few architectures aim for biological fidelity at the perceptual level. The only elaborate model of biologically-plausible sensory integration is demonstrated using a brain-based device (BBD) architecture. The embodied neural model called Darwin XI is constructed to investigate the integration of multi-sensory information (from touch sensors, laser, camera and magnetic compass) and the formation of place activity in the hippocampus during maze navigation \cite{Fleischer2007}. The neural network of Darwin XI consists of approximately 80,000 neurons with 1.2 million synapses and simulates 50 neural areas. In a lesion study, the robustness of the system was demonstrated by removing one or several sensory inputs and remapping of the sensory neuronal units.

Generally, cognitive architectures largely ignore cross-modal interaction. The architectures, including biologically- and cognitively-oriented ones, typicaly pursue a modular approach when dealing with different sensory modalities. At the same time, many psychological and neuroimaging experiments conducted in the past decades suggest that sensory modalities mutually affect one another. For example, vision alters auditory processing (e.g. McGurk effect \cite{McGurk1976}) and vice versa \cite{Shimojo2001}. While, some biomimetic architectures, such as BBD mentioned above, may be capable of representing cross-modal effects, to the best of our knowledge, this problem has not been investigated yet. 

\section{Attention} \label{section_5_attention}
Perceptual attention plays an important role in human cognition, as it mediates the selection of relevant and filters out irrelevant information from the incoming sensory data. However, it would be wrong to think of attention as a monolithic structure that makes a decision about what to process next. Rather, the opposite may be true. There is ample evidence in favor of attention as set of mechanisms affecting both perceptual and cognitive processes \cite{Tsotsos2014}. Currently, visual attention remains the most studied form of attention, as there are no comprehensive frameworks for other sensory modalities. Since only a few architectures have rudimentary mechanisms for modulating auditory data (OMAR \cite{Deutsch1997a}, iCub \cite{Ruesch2008}, EPIC \cite{Kieras2016} and MIDAS \cite{Gore2009}), this section will be dedicated to visual attention.

\begin{figure*}
  \includegraphics[width=0.9\textwidth]{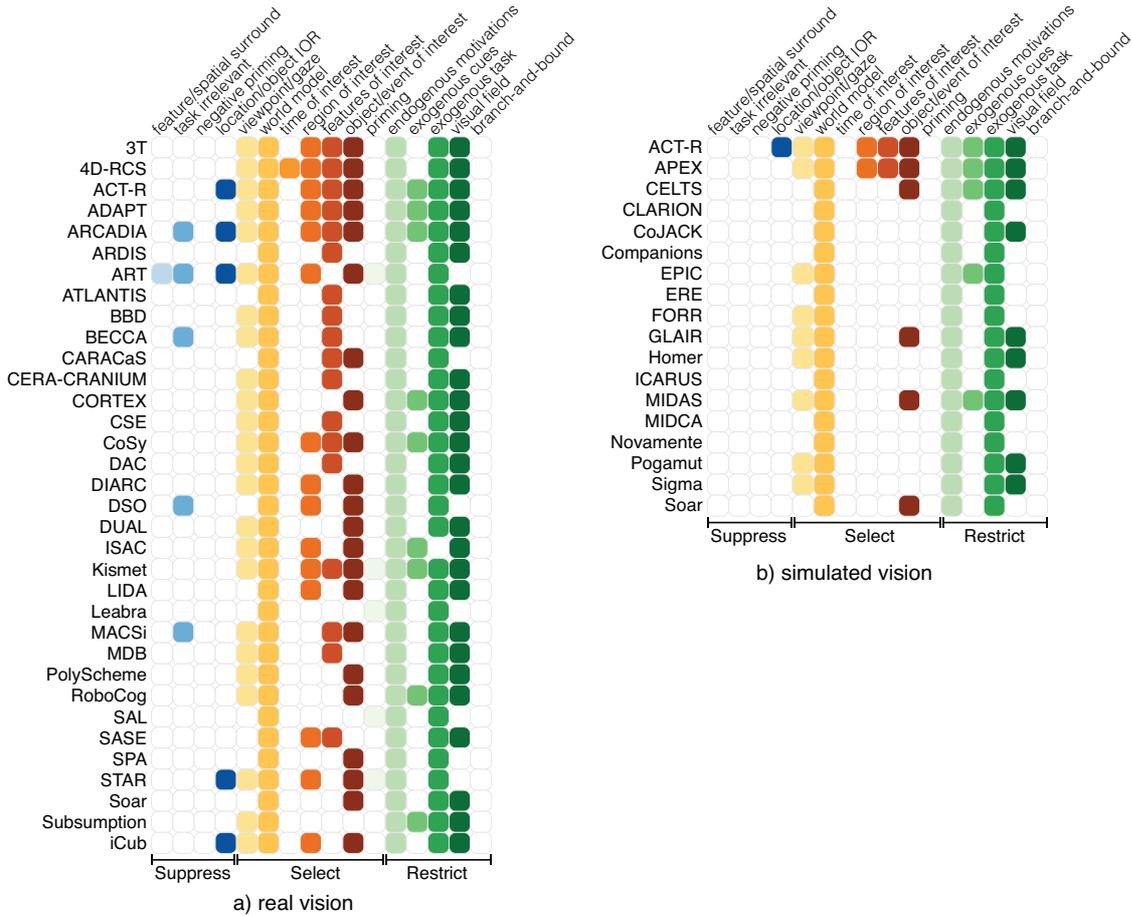}
\caption{A diagram showing visual attention mechanisms implemented in cognitive architectures for suppressing, selecting and restricting visual information. Architectures are ordered alphabetically. As in Section \ref{section_4_vision}, architectures in the left and right columns implement real and simulated vision respectively. Architectures not implementing vision or missing technical details are not shown. Suppression, selection and restriction mechanisms in the diagram are shown in shades of blue, yellow-red and green respectively.}
\label{fig_6_cog_arch_attention}       
\end{figure*}

For the following analysis, we use the taxonomy of attentional mechanisms proposed by Tsotsos \cite{Tsotsos2011}. Elements of attention are grouped into three classes of information reduction mechanisms: selection (choose one from many), restriction (choose some from many) and suppression (suppress some from many). Selection mechanisms include gaze and viewpoint selection, world model (selection of objects/events to focus on) and time/region/features/objects/events of interest. Restriction mechanisms serve to prune the search space by priming\footnote{Here we consider only priming in vision, other types of priming and their importance for learning are discussed in Section \ref{section_priming}} (preparing the visual system for input based on task demands), endogenous motivations (domain knowledge), exogenous cues (external stimuli), exogenous task (restrict attention to objects relevant for the task), and visual field (limited field of view). Suppression mechanisms consist of feature/spatial surround inhibition (temporary suppression of the features around the object while attending), task irrelevant stimuli suppression, negative priming, and location/object inhibition of return (a bias against returning attention to previous attended location or stimuli). Finally, branch-and-bound mechanisms combine elements of suppression, selection and restriction. For more detailed explanations of the attentional mechanisms and review of the relevant psychological and neuroscience literature refer to \cite{Tsotsos2011}.

The diagram in Figure \ref{fig_6_cog_arch_attention} shows a summary of the implemented visual attention elements in the cognitive architectures. Here we only consider the architectures with implemented real or simulated vision and omit projects for which not enough technical details are provided (same as in the previous section). It is evident that most of the implemented mechanisms of attention belong to the selection and restriction group. 

Only a handful of architectures implement suppression mechanisms. For instance, suppression of task irrelevant visual information is done in only three architectures: in BECCA irrelevant features are suppressed through the WTA mechanisms \cite{Rohrer2011a}, in DSO background features are suppressed to speed up processing \cite{Xiao2015}, in MACSi depth is used to ignore regions that are not reachable by the robot \cite{Lyubova2013} and in ARCADIA non-central regions of the visual field are inhibited at each cycle since the cue always appears in the center \cite{Bridewell2015}. Another suppression mechanism is inhibition of return (IOR), which prevents the visual system from attending to the same salient stimuli. In ACT-R activation values and distance are used to ignore objects for consecutive WHERE requests \cite{Nyamsuren2013a}. In ART, iCub and STAR an additional map is used to keep records of the attended locations. The temporal nature of inhibition can be modeled with a time decay function, as it is done in iCub \cite{Ruesch2008}. ARCADIA mentions a covert inhibition mechanism, however, the implementation details are not provided \cite{Bridewell2015}. 

Selection mechanisms are more commonly found in the cognitive architectures. For example, a world model by default is a part of any visual system. Viewpoint/gaze selection is a necessary component of active vision systems. Gaze control allows to focus on a region in the environment and viewpoint selection allows to get more information about the region/object by changing the distance to it or by viewing it from different angles. The physically embodied architectures automatically support viewpoint selection because a camera installed on a robot can be moved around the environment. Eye movements are usually simulated, but it is also possible to implement them on a humanoid robot (iCub, Kismet). 

Other mechanisms in this group include selection of various features/objects of interest. The selection of visual data to attend can be data-driven (bottom-up) or task-driven (top-down). The bottom-up attentional mechanisms identify salient regions whose visual features are distinct from the surrounding image features, usually along a combination of dimensions, such as color channels, edges, motion, etc. Some architectures resort to the classical visual saliency algorithms, such as Guided Search \cite{Wolfe1994} used in ACT-R \cite{Nyamsuren2013a} and Kismet \cite{Breazeal1999}, the Itti-Koch-Niebur model \cite{Itti1998} used by ARCADIA \cite{Bridewell2015}, iCub \cite{Ruesch2008} and DAC \cite{Mathews2012} or AIM \cite{Bruce2005} used in STAR \cite{Kotseruba2016}. Other approaches include filtering (DSO \cite{Xiao2015}), finding unusual motion patterns (MACsi \cite{Oudeyer}) or discrepancies between observed and expected data (RCS \cite{Albus2002}). 

 Top-down selection can be applied to further limit the sensory data provided by the bottom-up processing. For example, in visual search, knowing desired features of the object (e.g., the color red) can narrow down the options provided by the data-driven figure-ground segmentation. Many architectures resort to this mechanism to improve search efficiency (ACT-R \cite{Salvucci2000}, APEX \cite{Freed1998a}, ARCADIA \cite{Bello2016}, CERA-CRANIUM \cite{Arrabales2009b}, CHARISMA \cite{Conforth2011a}, DAC \cite{Mathews2012}). Another option is to use a hard-coded or learned heuristics. For example, CHREST looks at typical positions on a chess board \cite{Lane2009} and MIDAS replicates common eye scan patterns of pilots \cite{Gore2009}. The limitation of the current top-down approaches is that they can direct vision for only a limited set of predefined visual tasks, however, ongoing research in STAR attempts to address this problem \cite{Tsotsos2014,Tsotsos2011}. 
 
Restriction mechanisms allow to further reduce the complexity of vision by limiting the search space to certain features (priming), events (exogenous cues), space (visual field), objects (exogenous task) and knowledge (endogenous motivations). Exogenous task and endogenous motivations are implemented in all the architectures by default as domain knowledge and task instructions. In addition to that, attention can be modulated by internal signals such as motivation, emotions, which are discussed in the next section on action selection.

A limited visual field is a feature of any physically embodied system since most cameras do not have a 360-degree field of view. This feature is also present in some simulated vision systems. An ability to react to the sudden events (exogenous cues), such as fast movement or bright color, is common in social robotics applications (e.g. ISAC \cite{Kawamura2004}, Kismet \cite{Breazeal2000c}). On the other hand, priming allows to bias the visual system towards particular types of stimuli via task instruction. For example, by assigning more weight to skin-colored features during the saliency computation to improve human detection (Kismet \cite{Breazeal2001b}) or by utilizing the knowledge of the probable location of some objects to spatially bias detection (STAR \cite{Kotseruba2016}). Neural mechanisms of top-down priming are studied in ART models \cite{Carpenter1987b}.

Overall, visual attention is largely overlooked in cognitive architectures research with the exception of the biologically plausible visual models (e.g. ART) and the architectures that specifically focus on vision research (ARCADIA, STAR\footnote{Note that biologically plausible models of vision, such as Selective Tuning \cite{Tsotsos1995}, extensions to HMAX \cite{Walther2002,Walther2007,Riesenhuber2005} and others \cite{Frintrop2010}, support many of the listed attention mechanisms. However, embedding these models in the cognitive architecture is non-trivial. For instance, STAR represents ongoing work on integrating general purpose vision, as represented by the Selective Tuning model, with other higher-order cognitive processes.}). This is surprising because strong theoretical arguments as to its importance in dealing with the computational complexity of visual processing have been known for decades \cite{Tsotsos1990}.  Often the attention mechanisms found in the cognitive architectures are side-effects of other design decisions. For example, task constraints, world model and domain knowledge are necessary for the functioning of other aspects of intelligent system and are implemented by default. Limited visual field and viewpoint changes often result from physical embodiment. Otherwise, region of interest selection and visual reaction to exogenous cues are the two most common mechanisms explicitly included for optimizing visual processing. 

In the cognitive and psychological literature, attention is also used as a broad term for the allocation of limited resources \cite{Rosenbloom2015b}. For instance, in the Global Workspace Theory (GWT) \cite{Baars2005} attentional mechanisms are central to perception, cognition and action. According to GWT, the nervous system is organized as multiple specialized processes running in parallel. Coalitions of these processes compete for attention in the global workspace and the contents of the winning coalition are broadcast to all other processes. For instance, the LIDA architecture is an implementation of GWT\footnote{ In GWT focus of attention is associated with consciousness, however, this topic is beyond the scope of this survey.
} \cite{Franklin2012}. Other architectures influenced by the GWT include ARCADIA \cite{Bridewell2015} and CERA-CRANIUM \cite{Arrabales2009}.
	
Along the same lines, cognition is simulated as a set of autonomous independent processes in ASMO (inspired by Society of Mind \cite{Minsky1986}). Here an attention value for each process is either assigned directly or learned from experience. Attention values vary dynamically and affect action selection and resource allocation \cite{Novianto2010}. A similar idea is implemented in COGNET \cite{Purcell1993}, DUAL \cite{Kiryazov2007} and Copycat/Metacat \cite{Mitchell1990b} in which multiple processes also compete for attention. Other computational mechanisms, such as queue (PolyScheme \cite{Kurup2011}), Hopfield nets (CogPrime \cite{Ikle2011}) and modulators (MicroPsi \cite{Bach2015}) have been used to implement a focus of attention. In MLECOG, in addition to saccades that operate on visual data, mental saccades allow to switch between the most activated memory neurons \cite{Starzyk2015}.

\section{Action selection} \label{sec_action_selection}

Informally speaking, action selection determines at any point in time "what to do next". At the highest level, action selection can be split into the what part involving the decision making and the how part related to motor control \cite{Ozturk2009}. However, this distinction is not always explicitly made in the literature, where action selection may refer to goal, task or motor command. For example, in the MIDAS architecture action selection involves both the selection of the next goal to pursue and of actions that implement it \cite{Tyler1998}. Similarly, in MIDCA the next action is normally selected from a planned sequence if one exists. In addition to this, a different mechanism is responsible for the adoption of the new goals based on dynamically determined priorities \cite{Paisner2013}. In COGNET and DIARC, selection of a task/goal triggers an execution of the associated procedural knowledge \cite{Systems2014,Brick2007}. In DSO, the selector module chooses between available actions or decisions to reach current goals and sub-goals \cite{Ng2012a}. Since the treatment of action selection in various cognitive architectures is inconsistent, in the following discussion the action selection mechanisms may apply both to decision-making and motor actions. 

\begin{figure*}
  \includegraphics[width=0.9\textwidth]{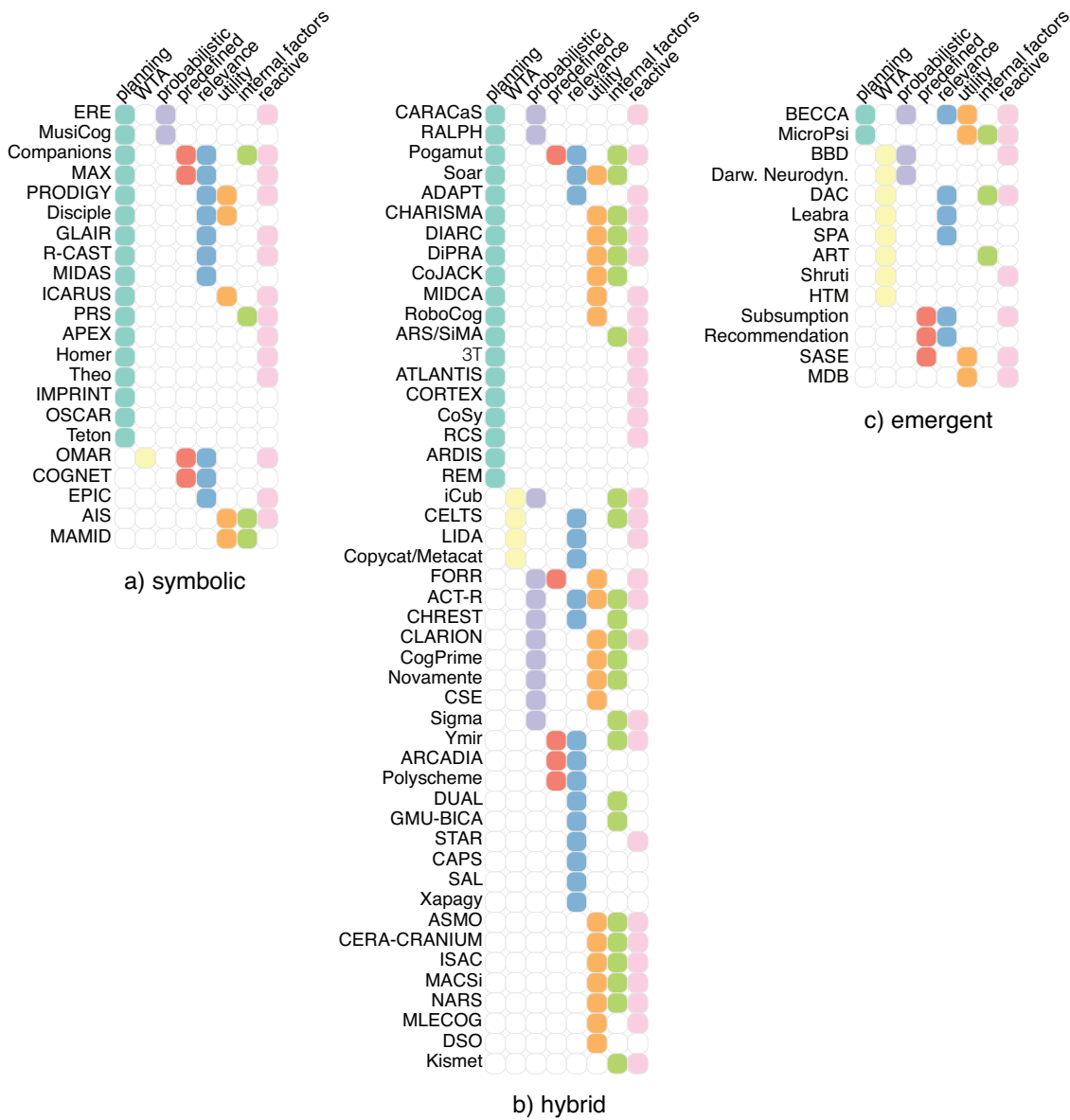}
\caption{A diagram of mechanisms involved in action selection. The visualization is organized in three columns grouping symbolic, hybrid and emergent architectures. Note that in this diagram (as well as in the diagrams for the sections \ref{section_7_memory} and \ref{section_8_learning}) the sorting order emphasizes clusters of architectures with similar action selection mechanisms (or memory and learning approaches respectively). Since the discussion follows the order in which different mechanisms are shown in the diagram, it can help to identify groups of architectures focusing on a particular mechanism which correspond to clusters within each column. A version of the diagram with an alphabetical ordering of the architectures is also available on the \href{http://jtl.lassonde.yorku.ca/project/cognitive_architectures_survey/action_selection.html}{project website}.}
\label{fig_7_cog_arch_action_selection}       
\end{figure*}

The diagram in Figure \ref{fig_7_cog_arch_action_selection} illustrates all implemented action selection mechanisms organized by the type of the corresponding architecture (symbolic, hybrid and emergent). We distinguish between two major approaches to action selection: planning and dynamic action selection. Planning refers to the traditional AI algorithms for determining a sequence of steps to reach a certain goal or to solve a problem in advance of their execution. In dynamic action selection, one best action is chosen among the alternatives based on the knowledge available at the time. For this category, we consider the type of selection (winner-take-all (WTA), probabilistic, predefined) and criteria for selection (relevance, utility, emotion). The default choice is always the best action based on the defined criteria (e.g. action with the highest activation value). Reactive actions are executed bypassing action selection. Finally, learning can also affect action selection but will be discussed in Section \ref{section_8_learning}. Note that these action selection mechanisms are not mutually exclusive and most of the architectures have more than one. Even though few architectures implement the same set of action selection mechanisms (as can be seen in the Figure \ref{fig_7_cog_arch_action_selection}), the whole space of valid combinations is likely much larger. Below we discuss approaches to action selection and what cognitive processes they correspond to in more detail.
\subsection{Planning vs reactive actions}
Predictably, planning is more common in the symbolic architectures, but can also be found in some hybrid and even emergent (MicroPsi \cite{Bach2015}) architectures. In particular, task decomposition, where the goal is recursively decomposed into subgoals, is a very common form of planning (e.g. Soar \cite{Derbinsky2012c}, Teton \cite{VanLehn1989}, PRODIGY \cite{Fink2005}). Other types of planning are also used: temporal (Homer \cite{Vere1991}), continual (CoSy \cite{Hawes2010a}), hierarchical task network (PRS \cite{DInverno2004}), generative (REM \cite{Murdock2008}), search-based (Theo \cite{Mitchell1990}), hill-climbing (MicroPsi \cite{Bach2015}), etc
     
Very few systems in our selection rely on classical planning alone, namely OSCAR, used for logical inference, and IMPRINT, which employs task decomposition for modeling human performance. Otherwise, planning is usually augmented with more dynamic action selection mechanisms to improve adaptability to the changing environment.

On the other end of the spectrum are reactive actions. These are executed immediately, suspending any ongoing activity and bypassing reasoning, similar to reflex actions in humans as automatic responses to stimuli. As demonstrated by the Subsumption architecture, combining multiple reactive actions can give rise to complex behaviors \cite{Brooks1990}, however, purely reactive systems are rare and reactive actions are used only under certain conditions. For example, they are used to protect the robot from the collision (ATLANTIS \cite{Gat1992}, 3T \cite{Aumann2001}) or to automatically respond to unexpected stimuli, such as fast moving objects or loud sounds (ISAC \cite{Kawamura2008}, Kismet \cite{Breazeal2000e}, iCub \cite{Ruesch2008}).  

\subsection{Dynamic action selection}
Dynamic action selection offers the most flexibility and can be used to model many phenomena typical for human and animal behavior. 

\textbf{Winner-take-all} is a selection process in neuronal networks where the strongest set of inputs is amplified, while the output from others is inhibited. It is believed to be a part of cortical processes and is often used in the computational models of the brain to make a selection from a set of decisions depending on the input. WTA and its variants are common in the emergent architectures such as HTM \cite{Byrne2015}, ART \cite{Carpenter2016}, SPA \cite{Eliasmith2015a}, Leabra \cite{OReilly2012}, DAC \cite{Verschure2003}, Darwinian Neurodynamics \cite{Fedor2017} and Shruti \cite{Wendelken2005}. Similar mechanisms are also used to find the most appropriate action in the architectures, where behavior emerges as a result of cooperation and competition of multiple processes running in parallel (e.g. Copycat/Metacat \cite{Marshall2002}, CELTS \cite{Nkambou2013}, LIDA \cite{Franklin2016}).

A \textbf{predefined order} of action selection may serve different purposes. For example, in the Subsumption architecture robot behavior is represented by a hierarchy of sub-behaviors, where higher-level behaviors override (subsume) the output of lower-level behaviors \cite{Brooks1990}. In FORR, the decision-making component considers options from advisors in the order of increasing expertise to achieve robust and human-like learning \cite{Epstein2006}. In Ymir priority is given to processes within the reactive layer first, followed by the content layer and then by the process control layer. Here the purpose is to provide a smooth real-time behavior generation. Each layer has a different upper bound on the perception-action time, thus reactive modules provide automatic feedback to the user (changing facial expressions, automatic utterances) while deliberative modules generate more complex behaviors \cite{Thorisson1998}.

The remaining action selection mechanisms include \textbf{finite-state machines} (FSM), which are frequently used to represent sequences of motor actions (ATLANTIS \cite{Gat1992}, CARACaS \cite{Huntsberger2011c}, Subsumption \cite{Brooks1990}) and even to encode the entire behavior of the system (ARDIS \cite{Martin2011}, STAR \cite{Kotseruba2016}). \textbf{Probabilistic action selection} is also common (iCub \cite{Vernon2010a}, CSE \cite{Henderson2013}, CogPrime \cite{Goertzel2010b}, Sigma \cite{Ustun2015a}, Darwinian Neurodynamics \cite{Fedor2017}, ERE \cite{Researce1992}, Novamente \cite{Goertzel2008b}).

\subsubsection{Action selection criteria}
There are several criteria that can be taken into account when selecting the next action: relevance, utility and affect (which includes motivations, affective states, emotions, moods, drives, etc.).

\textbf{Relevance} reflects how well the action corresponds to the current situation. This mainly applies to systems with symbolic reasoning and involves checking pre- and/or post-conditions of the rule before applying it (MAX \cite{Kuokka1991}, Disciple \cite{Tecuci2010}, EPIC \cite{Kieras2012a}, GLAIR \cite{Shapiro2010}, PRODIGY \cite{Veloso1998}, MIDAS \cite{Tyler1998}, R-CAST \cite{Fan2010}, Disciple \cite{Tecuci2010}, Companions \cite{Hinrichs2007}, Ymir \cite{Thorisson1998}, Pogamut \cite{Brom2007}, Soar \cite{Laird2012}, ACT-R \cite{Lebiere2013}).

\textbf{Utility} of the action is a measure of its expected contribution to achieving the current goal (CERA-CRANIUM \cite{Arrabales2009}, CHARISMA \cite{Conforth2011a}, DIARC \cite{Brick2007}, MACSi \cite{Oudeyer}, MAMID \cite{Hudlicka2006}, NARS \cite{Wang2015a}, Novamente \cite{Goertzel2008b}). Some architectures also perform a "dry run" of candidate actions and observe their effect to determine their utility (MLECOG \cite{Starzyk2015}, RoboCog \cite{Bandera2014}). Utility can also take into account the performance of the action in the past and improve the behavior in the future via reinforcement learning (ASMO \cite{Novianto2014a}, BECCA \cite{Rohrer2011b}, CLARION \cite{Sun2002}, PRODIGY \cite{Carbonell1992}, CSE \cite{Henderson2013}, CoJACK \cite{Ritter2009}, DiPRA \cite{Pezzulo2009}, Disciple \cite{Tecuci2010}, DSO \cite{Ng2012a}, FORR \cite{Epstein2006}, ICARUS \cite{Shapiro2001a}, ISAC \cite{Kawamura2008}, MDB \cite{Bellas2005}, MicroPsi \cite{Bach2011}, Soar \cite{Laird2012}). Other machine learning techniques are also used to associate goals with successful behaviors in the past (MIDCA \cite{Paisner2013}, SASE \cite{Weng2006}, CogPrime \cite{Goertzel2010b}). 

Finally, \textbf{internal factors} do not determine the next behavior directly, but rather bias the selection. For simplicity, we will consider short-term, long-term and lifelong factors corresponding to emotions, drives and personality traits in humans. Given the impact of these factors on human decision making and other cognitive abilities it is important to model emotion and affect in cognitive architectures, especially in the areas of human-computer interaction, social robotics and virtual agents. Not surprisingly, this is where most of the effort has been spent so far (see Section \ref{subsection_virtual_agents} on applications of intelligent agents with realistic personalities and emotions).

\textbf{Artificial emotions} in cognitive architectures are typically modeled as transient states (associated with anger, fear, joy, etc.) that influence cognitive abilities \cite{Hudlicka2004,Goertzel2008c,Wilson2013}. For instance, in CoJACK morale and fear emotions can modify plan selection. As a result, plans that confront a threat have higher utility when morale is high, but lower utility when fear is high \cite{Ritter2009}. Other examples include models of stress affecting decision-making (MAMID \cite{Conference2001}), emotions of joy/sadness affecting blackjack strategy (CHREST \cite{Schiller2012}), analogical reasoning in the state of anxiety (DUAL \cite{Feldman2009a}), effect of arousal on memory (ACT-R \cite{Cochran2006}), positive and negative affect depending on goal satisfaction (DIARC \cite{Scheutz2009}) and emotional learning in HCI scenario (CELTS \cite{Nkambou2013}). The largest attempt at implementing a theory of appraisal led to the creation of Soar-Emote, a model based on the ideas of Damasio and Gratch \& Marsella \cite{Marinier2004}.
 
\textbf{Drives} are another source of internal motivations. Broadly speaking, they represent the basic physiological needs such as food and safety, but can also include high-level or social drives. Typically, any drive has strength that diminishes as the drive is satisfied. As the agent works towards its goals, several drives may affect its behavior simultaneously. For example, in ASMO three relatively simple drives, liking color red, praise and happiness of the user and the robot, bias action selection by modifying the weights of the corresponding modules \cite{Novianto2014a}.  In CHARISMA preservation drives (avoid harm and starvation) combined with curiosity and desire for self-improvement guide behavior generation \cite{Conforth2011}. In MACSi curiosity drives exploration towards areas where the agent learns the fastest \cite{Nguyen2013}. Likewise, in CERA-CRANIUM curiosity, fear and anger affect exploration of the environment by the mobile robot \cite{Moreno2006}. The behavior of social robot Kismet is organized around satiating three homeostatic drives: to engage people, to engage toys, and to rest. These drives and external events, contribute to the affective state (or "mood") of the robot and its expression as anger, disgust, fear, joy, sorrow and surprise via facial gestures, stance or changes in the tone of its voice. Although its motivational mechanisms are hard-coded, Kismet demonstrates one of the widest repertoire of emotional behaviors of all the cognitive architectures we reviewed \cite{Breazeal2003}.

Unlike emotions, which have a transient nature, \textbf{personality traits} are unique long-term patterns of behavior that manifest themselves as consistent preferences in terms of internal motivations, emotions, decision-making, etc. Most of the identified personality traits can be reduced to a small set of dimensions/factors necessary and sufficient for broadly describing human personality (e.g. the Five-Factor Model (FFM) \cite{mccrae1992introduction} and Mayers-Briggs model \cite{myers1998mbti}). Likewise, personality in cognitive architectures is often represented by several factors/dimensions, not necessarily based on a known model. These traits, in turn, are related to the emotions and/or drives that the system is likely to experience.

In the simplest case a single parameter suffices to create a systematic bias in the behavior of the system. Both NARS \cite{Wang2006} and CogPrime \cite{Goertzel2013b} use the "personality parameter" to define how much evidence is required to evaluate the truth of the logical statement or to plan the next action respectively. The larger the value of the parameter, the more "conservative" the system appears to be. 

In Novamente personality traits of a virtual animal (e.g. aggressiveness, curiosity, playfullness) are linked to emotional states and actions via probabilistic rules \cite{Goertzel2008c}. Similarly, in AIS traits (e.g. nasty, phlegmatic, shy, confident, lazy) are assigned an integer value that defines the degree to which they are exhibited. Abstract rules define what behaviors are more likely given the personality \cite{Rousseau1997a}. In CLARION personality type determines the baseline strength and initial deficit (i.e. proclivity towards) of the many predefined drives. The mapping is embodied in a pretrained neural network \cite{Sun2016a}.

Remarkably, even these simple models can generate a spectrum of personalities. For instance, the Pogamut agent has 9 possible states and 5 personality factors (based on FFM) resulting in 45 distinct mappings. Each of these mappings can produce any of the 12 predefined intentions to a varying degree and generate a wide range of behaviors \cite{Rosenthal2012}.

CLARION and MAMID deserve a special mention. CLARION provides a cognitively-plausible framework that is capable of addressing emotions, drives and personality traits and relating them to other cognitive processes including decision-making. Three aspects of emotion are modeled: reactive affect (or unconscious experience of emotion), deliberative appraisal (potentially conscious) and coping/action (following appraisal). Thus emotion emerges as an interaction between the explicit and implicit processes and involves (and affects) perception, action and cognition \cite{Sun2016,Sun2016a}. Several CLARION models have been validated using psychological data, computational model of the FFM personalities \cite{Sun2011}, coping with bullying at school \cite{Wilson2014b}, performance degradation under pressure \cite{Wilson2009} and model of stereotype bias induced by social anxiety \cite{Wilson2010}.  

MAMID is a model of both generation and effect of emotions resulting from external events, internal interpretations, goals and personality traits. Internally, belief nets relate the task- and individual-specific criteria, e.g. whether a goal failure will result in anxiety in a particular agent \cite{Hudlicka2008a}. MAMID has been instantiated in two domains: peacekeeping mission training \cite{Hudlicka2001} and search-and-rescue tasks \cite{Hudlicka2005}. Of great value are also contributions towards developing an interdisciplinary theory of emotions and design guidelines \cite{Hudlicka2016,Reisenzein2013}.

\section{Memory} \label{section_7_memory}

Memory is an essential part of any systems-level cognitive model, regardless of whether the model is being used for studying the human mind or for solving engineering problems. Thus, nearly all the architectures featured in this review have memory systems that store intermediate results of computations, enabling learning and adaptation to the changing environment. However, despite their functional similarity, the particular implementations of memory systems differ significantly and depend on the research goals and conceptual limitations, such as biological plausibility and engineering factors (e.g. programming language, software architecture, use of frameworks, etc.). In the cognitive architecture literature, memory is described in terms of its duration (short- and long-term) and type (procedural, declarative, semantic, etc.), although it is not necessarily implemented as separate knowledge stores.

The multi-store memory model is influenced by the Atkinson-Shiffrin model (1968) \cite{Atkinson1968}, later modified by  Baddeley \cite{Baddeley1974}. This view of memory is dominant in psychology, but its utility for engineering is questioned by some because it does not provide a functional description of various memory mechanisms \cite{Perner2011a}. Nevertheless, most architectures do distinguish between various memory types, although the naming conventions differ depending on the conceptual background. For instance, the architectures designed for planning and problem solving have short- and long-term memory storage systems but do not use terminology from cognitive psychology. The long-term knowledge in planners is usually referred to as a knowledge base for facts and problem-solving rules, which correspond to semantic and procedural long-term memory (e.g. Disciple \cite{Tecuci2000}, MACsi \cite{Oudeyer}, PRS \cite{Georgeff1986}, ARDIS \cite{Martin2011}, ATLANTIS \cite{Gat1992}, IMPRINT \cite{Brett2002}). Some architectures also save previously implemented tasks and solved problems, imitating episodic memory (REM \cite{Murdock2001},  PRODIGY \cite{Veloso1994}). The short-term storage in planners  is usually represented by a current world model or the contents of the goal stack.

\begin{figure*}
  \includegraphics[width=0.85\textwidth]{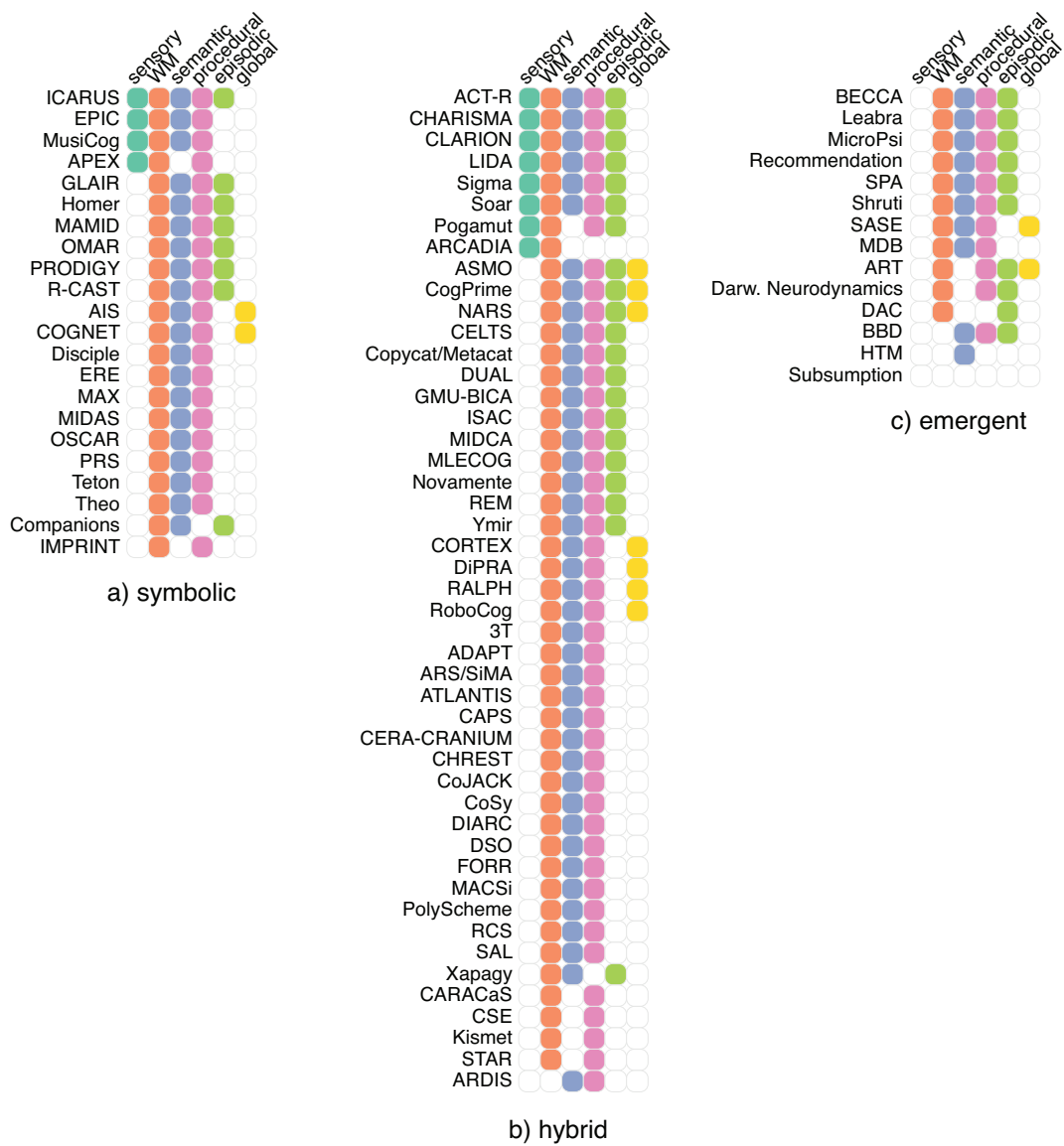}
\caption{A diagram showing types of memory implemented in the cognitive architectures. The symbolic, hybrid and emergent architectures are shown in separate columns. The architectures where memory system is unified, i.e. no distinction is made between short-, long-term and other types of memory, are labeled as "global". A version of the diagram with an alphabetical ordering of the architectures is also available on the \href{http://jtl.lassonde.yorku.ca/project/cognitive_architectures_survey/memory.html}{project website}.}
\label{fig_8_cog_arch_memory}       
\end{figure*}

 Figure \ref{fig_8_cog_arch_memory} shows a visualization of various types of memory implemented by the architectures. Here we follow the convention of distinguishing between the long-term and short-term storage. Long-term storage is further subdivided into semantic, procedural and episodic types, which store factual knowledge, information on what actions should be taken under certain conditions respectively and episodes from the personal experience of the system respectively. Short-term storage is split into sensory and working memory following \cite{Cowan2008}. Sensory or perceptual memory is a very short-term buffer that stores several recent percepts. Working memory is a temporary storage for percepts that also contains other items related to the current task and is frequently associated with the current focus of attention. 

\subsection{Sensory memory}
The purpose of sensory memory is to cache the incoming sensory data and preprocess it before transferring it to other memory structures. For example, iconic memory assists in solving continuity and maintenance problems, i.e. identifying separate instances of the objects as being the same and retaining the impression of the object when unattended (ARCADIA \cite{Bridewell2015}). Similarly, echoic memory allows an acoustic stimulus to persist long enough for perceptual binding and feature extraction, such as pitch extraction and grouping (MusiCog \cite{Maxwell2014a}). The decay rate for items in sensory memory is believed to be tens of milliseconds (EPIC \cite{Kieras2014b}, LIDA \cite{Baars2008}) for visual data and longer for audio data (MusiCog \cite{Maxwell2014a}), although the time limit is not always specified. Other architectures implementing this memory type include Soar \cite{Wintermute2009c}, Sigma \cite{Rosenbloom2015a}, ACT-R \cite{Nyamsuren2013a}, CHARISMA \cite{Conforth2012}, CLARION \cite{Sun2012}, ICARUS \cite{Langley2005a} and Pogamut \cite{Brom2007}.

\subsection{Working memory}
\label{subsection_working_memory}
Working memory can be defined as a mechanism for temporary storage of information related to the current task. It is critical for cognitive capacities such as attention, reasoning and learning, thus every cognitive architecture in our list implements it in some form. Particular realizations of working memory differ mainly in what information is being stored, how it is represented, accessed and maintained. Furthermore, some cognitive architectures contribute to ongoing research about the processes involved in encoding, manipulation and maintenance of information in human working memory and its relationship to other processes in the human brain. 
Despite the importance of working memory for human cognition, relatively few publications provide sufficient detail about its internal organization and connections to other modules. Often it is summarized in a few words, e.g. "current world state" or "data from the sensors". Based on these statements we infer that in many architectures working memory or an equivalent structure serves as a cache for the current world model, the state of the system and/or current goals. Although there are no apparent limitations on the capacity of working memory, new goals or new sensory data usually overwrite the existing content. This simplified account of working memory can be found in many symbolic architectures (3T \cite{PeterBonasso1997}, ATLANTIS \cite{Gat1992}, Homer \cite{Vere1990}, IMPRINT \cite{Brett2002}, MIDCA \cite{Cox2012}, PRODIGY \cite{Veloso1994c}, R-CAST \cite{Jajodia2010}, RALPH \cite{Ogasawara1991}, REM \cite{Murdock2008}, Theo \cite{Mitchell1990}, Disciple \cite{Tecuci2007b}, OSCAR \cite{Pollock1993a}, ARS/SiMA \cite{Schaat2014}, CoJACK \cite{Ritter2012}, ERE \cite{Bresina1990}, MAX \cite{Kuokka1989}, Companions \cite{Forbus2009a}, RCS \cite{Albus2002a}) as well as hybrids (CSE \cite{Henderson2013}, MusiCog \cite{Maxwell2014a}, MicroPsi \cite{Bach2013}, Pogamut \cite{Brom2007}). 
More cognitively plausible models of working memory use an activation mechanism. Some of the earliest models of activation of working memory contents were implemented in ACT-R. As before,  working memory holds the most relevant knowledge, which is retrieved from the long-term memory as determined by bias. This bias is referred to as activation and consists of base-level activation (which may decrease or increase) with every access and may also include the spreading activation from neighboring elements. The higher the activation of the element, the more likely it is to enter working memory and directly affect the behavior of the system \cite{Anderson1996}. This applies to the graph-based knowledge representation where nodes refer to concepts and weights assigned to edges correspond to the associations between the concepts (Soar \cite{Laird2014}, CAPS \cite{Sanner1999}, ADAPT \cite{Benjamin2004}, DUAL \cite{Kokinov1996}, Sigma \cite{Pynadath2014}, CELTS \cite{Nkambou2013}, NARS \cite{Wang2007b}, Novamente \cite{Goertzel2007b}, MAMID \cite{Hudlicka2010}). Naturally, activation can be used in neural network representation as well (Shruti \cite{Wendelken2005}, CogPrime \cite{Ikle2011}, Recommendation Architecture \cite{Coward2011}, SASE \cite{Weng2010}, Darwinian Neurodynamics \cite{Fedor2017}). Activation mechanisms contribute to modeling many properties of working memory, such as limited capacity, temporal decay, rapid updating as the circumstances change, connection to other memory components and decision-making.
Another common paradigm, the blackboard architecture, represents memory as a shared repository of goals, problems and partial results which can be accessed and modified by the modules running in parallel (AIS \cite{Hayes-Roth1996}, CERA-CRANIUM \cite{Arrabales2011}, CoSy \cite{Sjoo2010}, FORR \cite{Epstein1992a}, Ymir \cite{Thorisson1999a}, LIDA \cite{Franklin2016}, ARCADIA \cite{Bello2016}, Copycat/Metacat \cite{Marshall2006}, CHARISMA \cite{Conforth2012}, PolyScheme \cite{Cassimatis2007b}, PRS \cite{Georgeff1998}). The solution to the problem is obtained by continuously updating the shared short-term storage with information from specialized heterogeneous modules analogous to a group of people completing a jigsaw puzzle. Although not directly biologically plausible, this paradigm works well in situations where many disparate sources of information must be combined to solve a complex real-time task.
Finally, neuronal models of working memory based on the biology of the prefrontal cortex are realized within SPA \cite{Stewart2014a}, ART \cite{Grossberg2007} and Leabra \cite{OReilly2006a}. They demonstrate a range of phenomena (validated on human data) such as adaptive switching between rapid updating of the memory and maintaining previously stored information \cite{OReilly2006} and reduced accuracy with time (e.g. in a list memory task) \cite{Stewart2014a}.
In some robotics architectures, ego-sphere (discussed in Section \ref{fig_4_cog_arch_perception}) besides enabling spatiotemporal integration also provides representation for the current location and the orientation of the robot within the environment (iCub \cite{Ruesch2008}, ISAC \cite{Kawamura2008}, MACSi \cite{Oudeyer}). Along the same lines, Kismet and RoboCog \cite{Bachiller2008} use a map built on the camera reference frame to maintain information about recently perceived regions of interest. 

Since, by definition, working memory is a relatively small temporary storage, for biological realism, its capacity should be limited.  However, there is no consensus on of how this should be done. For instance, in GLAIR the contents of working memory are discarded when the agent switches to a new problem \cite{Shapiro2010}. A more common approach is to gradually remove items from the memory based on their recency or relevance in the changing context. The CELTS architecture implements this principle by assigning an activation level to percepts that is proportional to the emotional valence of a perceived situation. This activation level changes over time and as soon as it falls below a set threshold, the percept is discarded \cite{Nkambou2013}. The Novamente Cognitive Engine has a similar mechanism, where atoms stay in memory as long as they build links to other memory elements and increase their utility \cite{Goertzel2007b}. It is still unclear if under these conditions the size of working memory can grow substantially without any additional restrictions. 

To prevent unlimited growth, a hard limit can be defined for the number of items in memory, for example, 3-6 objects in ARCADIA \cite{Bello2016}, 4 chunks in CHREST \cite{Lloyd-kelly} or up to 20 items in MDB \cite{Bellas2004}. Then as the new information arrives, the oldest or the most irrelevant items would be deleted to avoid overflow. Items can also be discarded if they have not been used for some time. The exact amount of time can vary from 4-9 seconds (EPIC \cite{Kieras2009a}) to 5 sec (MIDAS \cite{Hooey2010}, CERA-CRANIUM \cite{Arrabales2011}) to tens of seconds (LIDA \cite{Franklin2007}). In the Recommendation Architecture, a different approach is taken so that the limit of 3-4 items in working memory emerges naturally from the structure of the memory system, and not the external parameter setting \cite{Coward2011}. 

\subsection{Long-term memory}
Long-term memory (LTM) preserves a large amount of information for a very long time. Typically, it is divided into the procedural memory of implicit knowledge (e.g. motor skills and routine behaviors) and declarative memory, which contains (explicit) knowledge. The latter is further subdivided into semantic (factual) and episodic (autobiographical) memory. 

The dichotomies between the explicit/implicit and the declarative/procedural dimensions of long-term memories are usually merged. One of the few exceptions is CLARION, where procedural and declarative memories are separate and both subdivided into an implicit and explicit component. This distinction is preserved on the level of knowledge representation: implicit knowledge is captured by distributed sub-symbolic structures like neural networks, while explicit knowledge has a transparent symbolic representation \cite{Sun2012}.

Long-term memory is a storage for innate knowledge that enables operation of the system, therefore almost all architectures implement procedural and/or semantic memory. The procedural memory contains knowledge about how to get things done in the task domain. In symbolic production systems, procedural knowledge is represented by a set of if-then rules preprogrammed or learned for a particular domain (3T \cite{Firby1989}, 4CAPS \cite{Varma2007}, ACT-R \cite{Lebiere2013}, ARDIS \cite{Martin2011}, EPIC \cite{Kieras2004}, SAL \cite{Herd2014}, Soar \cite{Lindes2016}, APEX \cite{Freed2000}). Other variations include sensory-motor schemas (ADAPT \cite{Benjamin2004}), task schemas (ATLANTIS \cite{Gat1992}) and behavioral scripts (FORR \cite{Epstein1994b}). In emergent systems, procedural memory may contain sequences of state-action pairs (BECCA \cite{Rohrer2009}) or ANNs representing perceptual-motor associations (MDB \cite{Salgado2012}).

Semantic memory stores facts about the objects and relationships between them. In the architectures that support symbolic reasoning, semantic knowledge is typically implemented as a network-like ontology, where nodes correspond to concepts and links represent relationships between them (Disciple \cite{Boicu2003}, MIDAS \cite{Corker1997}, Soar \cite{Lindes2016}, CHREST \cite{Lloyd-kelly2015}). In emergent architectures factual knowledge is represented as patterns of activity within the network (BBD \cite{Krichmar2005a}, SHRUTI \cite{Shastri2007}, HTM \cite{Lavin2016}, ART \cite{Carpenter2001a}).

Episodic memory stores specific instances of past experience. These can later be reused if a similar situation arises (MAX \cite{Kuokka1991}, OMAR \cite{Deutsch2006}, iCub \cite{Vernon2010a}, Ymir \cite{Thorisson1999a}). However, these experiences can also be exploited for learning new semantic or procedural knowledge. For example, CLARION saves action-oriented experiences as "input, output, result" and uses them to bias future behavior \cite{Sun1998a}. Similarly, BECCA stores sequences of state-action pairs to make predictions and guide the selection of system actions \cite{Rohrer2009}, and MLECOG gradually builds 3D scene representation from perceived situations \cite{Jaszuk2016}. MAMID \cite{Hudlicka2002a} saves the past experience together with the specific affective connotations (positive or negative), which affect the likelihood of selecting similar actions in the future. Other examples include R-CAST \cite{Barnes2008}, Soar \cite{Nuxoll2007}, Novamente \cite{Lian2010} and Theo \cite{Mitchell1989a}.

\subsection{Global memory}
Despite the evidence for the distinct memory systems, some architectures do not have separate representations for different kinds of knowledge or short- vs long-term memory, and instead, use a unified structure to store all information in the system. For example, CORTEX and RoboCog use an integrated, dynamic multi-graph object which can represent both sensory data and high-level symbols describing the state of the robot and the environment \cite{Bandera,Bustos2013}. Similarly, AIS implements a global memory, which combines a knowledge database, intermediate reasoning results and the cognitive state of the system \cite{Washingtona1992}. DiPRA uses Fuzzy Cognitive Maps to represent goals and plans \cite{Pezzulo2007}. NARS represents all empirical knowledge, regardless of whether it is declarative, episodic or procedural, as formal sentences in Narcese \cite{Wang2015b}.  Similarly, in some emergent architectures, such as SASE \cite{Weng2010} and ART \cite{Carpenter2001a}, the role of neurons as working or long-term memory is dynamic and depends on whether the neuron is firing.

Overall, research on memory in the cognitive architectures mainly concerns its structure, representation and retrieval. Relatively little attention has been paid to the challenges associated with maintaining a large-scale memory store since both the domains and the time spans of the intelligent agents are typically limited. In comparison, early estimates of the capacity of the human long-term memory are within 1.5 gigabits or on the order of 100K concepts \cite{Landauer1986}, while more recent findings suggest that human brain capacity may be orders of magnitude higher \cite{Bartol2015}.  However, scaling na{\"i}ve implementations even to the lowest estimated size of human memory likely will not be possible despite the increase in available computational power. Thus alternative solutions include tapping into existing methods for large-scale data management and improving the retrieval algorithms. Both avenues have been explored, the former by Soar and ACT-R, which used PostgreSQL relational database to load concepts and relations from WordNet \cite{Derbinsky2012d,Douglass2009}, and the latter by the Companions architecture \cite{Forbus2016}. Alternatively, SPA supports a biologically plausible model of associative memory capable of representing over 100K concepts in WordNet using a network of spiking neurons \cite{Crawford2010}. 

\section{Learning} \label{section_8_learning}

\begin{figure*}
  \includegraphics[width=1.00\textwidth]{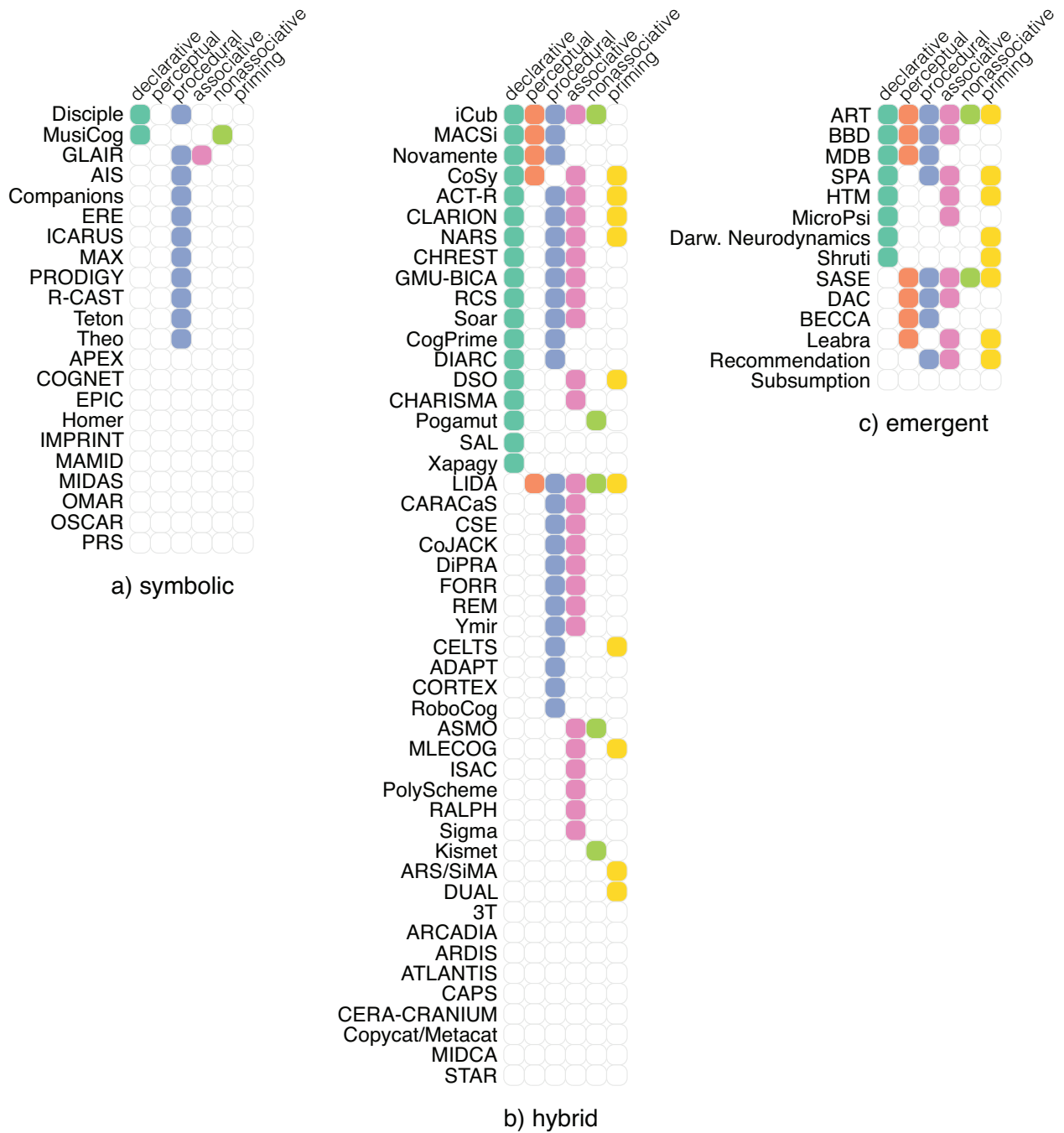}
\caption{A diagram summarizing types of learning implemented in the cognitive architectures. Symbolic, hybrid and emergent architectures are grouped and presented in different columns. The architectures are arranged according to the similarity between the implemented learning types. A version of the diagram with an alphabetical ordering of the architectures is also available on the \href{http://jtl.lassonde.yorku.ca/project/cognitive_architectures_survey/learning.html}{project website}.}
\label{fig_9_cog_arch_learning}       
\end{figure*}

Learning is the capability of a system to improve its performance over time. Ultimately, any kind of learning is based on experience. For example, a system may be able to infer facts and behaviors from the observed events or from results of its own actions. The type of learning and its realization depend on many factors, such as design paradigm (e.g. biological, psychological), application scenario, data structures and the algorithms used for implementing the architecture, etc. However, we will not attempt to analyze all these aspects given the diversity and number of the cognitive architectures surveyed. Besides, not all of these pieces of information can be easily found in the publications.
Thus, a more general summary is preferable, where types of learning are defined following taxonomy by Squire (\cite{Squire1992}). Learning is divided into declarative or explicit knowledge acquisition and non-declarative, which includes perceptual, procedural, associative and non-associative types of learning. Figure \ref{fig_9_cog_arch_learning} shows a visualization of  these types of learning for all cognitive architectures.

\subsection{Perceptual learning}
Although many systems use pre-learned components for processing perceptual data, such as object and face detectors or classifiers, we do not consider these here. Perceptual learning applies to the architectures that actively change the way sensory information is handled or how patterns are learned on-line. This kind of learning is frequently performed to obtain implicit knowledge about the environment, such as spatial maps (RCS \cite{Schlenoff2005}, AIS \cite{Hayes-Roth1993}, MicroPsi \cite{Bach2007}), clustering visual features (HTM \cite{Kostavelis2012a}, BECCA \cite{Rohrer2011}, Leabra \cite{OReilly2013}) or finding associations between percepts. The latter can be used within the same sensory modality as in the case of the agent controlled by Novamente engine, which selects a picture of the object it wants to get from a teacher \cite{Goertzel2008f}. Learning can also occur between different modalities, for instance, the robot based  on the SASE architecture learns the association between spoken command and action \cite{Weng2002a} and Darwin VII (BBD) learns to associate taste value of the blocks with their visual properties \cite{Edelman2007a}.

\subsection{Declarative learning}

Declarative knowledge is a collection of facts about the world and various relationships defined between them. In many production systems such as ACT-R and others, which implement chunking mechanisms (SAL \cite{Jilk2008}, CHREST \cite{Schiller2012}, CLARION \cite{Sun1999c}), new declarative knowledge is learned when a new chunk is added to declarative memory (e.g. when a goal is completed). Similar acquisition of knowledge has also been demonstrated in systems with distributed representations. For example, in DSO knowledge can be directly input by human experts or learned as contextual information extracted from labeled training data \cite{Ng2012a}. New symbolic knowledge can also be acquired by applying logical inference rules to already known facts (GMU-BICA \cite{Samsonovich2008a}, Disciple \cite{Boicu2005c}, NARS \cite{Wang2010}). In many biologically inspired systems learning new concepts usually takes the form of learning the correspondence between the visual features of the object and its name (iCub \cite{DiNuovo2014}, Leabra \cite{OReilly2013}, MACsi \cite{Oudeyer}, Novamente \cite{Goertzel2008f}, CoSy \cite{Hawes2010a}, DIARC \cite{Yu2010}). 

\subsection{Procedural learning}
Procedural learning refers to learning skills, which happens gradually through repetition until the skill becomes automatic. The simplest way of doing so is by accumulating examples of successfully solved problems to be reused later (e.g. AIS \cite{Hayes-Roth1993}, R-CAST \cite{Fan2010}, RoboCog \cite{Manso2014}). For instance, in a navigation task, a traversed path could be saved and used again to go between the same locations later (AIS \cite{Hayes-Roth1993}). Obviously, this type of learning is very limited and further processing of accumulated experience is needed to improve efficiency and flexibility. 

Explanation-based learning (EBL) \cite{Minton1989} is a common technique for learning from experience found in many architectures with symbolic representation for procedural knowledge (PRODIGY \cite{Etzioni1993}, Teton \cite{Vanlehn1989e}, Theo \cite{Vanlehn1989e}, Disciple \cite{Boicu2005c}, MAX \cite{Kuokka1991}, Soar \cite{Laird2012a}, Companions \cite{Friedman2010}, ADAPT \cite{Benjamin2010}, ERE \cite{Researce1992}, REM \cite{Murdock2008}, CELTS \cite{Faghihi2011d}, RCS \cite{Albus1995c}). In short, it allows generalization of the explanation of a single observed instance into a general rule. However, EBL does not extend the problem domain, but rather makes solving problems more efficient in similar situations. A known drawback of this technique is that it can result in too many rules (also known as the "utility problem"), which may slow down the inference. To avoid the explosion in the model knowledge, various heuristics can be applied, e.g. adding constraints on events that a rule can contain (CELTS \cite{Faghihi2011d}) or eliminating low-use chunks (Soar \cite{Kennedy2003}). Although EBL is not biologically inspired, it has been shown that in some cases human learning may exhibit EBL-like behavior \cite{Vanlehn1990}.

Symbolic procedural knowledge can also be obtained by inductive inference (Theo \cite{Mitchell1989a}, NARS \cite{Slam2015}), learning by analogy (Disciple \cite{Tecuci1990}, NARS \cite{Slam2015}), behavior debugging (MAX \cite{Kuokka1991}), probabilistic reasoning (CARACaS \cite{Huntsberger2011}), correlation and abduction (RCS \cite{Albus1995c}) and explicit rule extraction (CLARION \cite{Sun2011a}). 

\subsection{Associative learning}
Associative learning is a broad term for decision-making processes influenced by reward and punishment. In behavioral psychology, it is studied within two major paradigms: classical (Pavlovian) and instrumental (operant) conditioning. Reinforcement learning (RL) and its variants, such as temporal difference learning, Q-learning, Hebbian learning, etc., are commonly used in computational models of associative learning. Furthermore, there is substantial evidence that error-based learning is fundamental for decision-making and motor skill acquisition \cite{Holroyd2002,Niv2009,Seidler2013}.

The simplicity and efficiency of reinforcement learning make it one of the most common techniques with nearly half of all the cognitive architectures using it to implement associative learning. The advantage of RL is that it does not depend on the representation and can be used in the symbolic, emergent or hybrid architectures. One of the main uses of this technique is developing adaptive behavior. In systems with symbolic components it can be accomplished by changing the importance of actions and beliefs based on their success/failure (e.g. RCS \cite{Albus2005}, NARS \cite{Slam2015}, REM \cite{Murdock2008}, RALPH \cite{Ogasawara1993a}, CHREST \cite{Schiller2012}, FORR \cite{Gordon2011a}, ACT-R \cite{Cao2015}, CoJACK \cite{Ritter2009a}). 

In the hybrid and emergent systems, reinforcement learning can establish associations between states and actions. One application is sensorimotor reconstruction. The associations are often established in two stages: a "motor babbling" stage, during which the system performs random actions to accumulate data, followed by the learning stage where a model is built using the accrued experiences (ISAC \cite{Kawamura2008}, BECCA \cite{Rohrer2009}, iCub \cite{Metta2010}, DiPRA \cite{Pezzulo2009}, CSE \cite{Henderson2011}). Associative learning can be applied when analytical solutions are hard to obtain as in the case of soft-arm control (ISAC \cite{Kawamura2008}) or when smooth and natural looking behavior is desired (e.g. Segway platform control using BBD \cite{Krichmar2012}).

\subsection{Non-associative learning}
Non-associative learning, as the name suggests, does not require associations to link stimuli and responses together. Habituation and sensitization are commonly identified as two types of non-associative learning. Habituation describes gradual reduction in the strength of response to repeated stimuli. The opposite process occurs during sensitization, i.e. repeated exposure to stimuli causes increased response. Because of their simplicity these types of learning are considered a prerequisite for other forms of learning. For instance, habituation filters out irrelevant stimuli and helps to focus on important stimuli \cite{Rankin2009}, especially in the situations when positive or negative rewards are absent \cite{Weng2006}.

Most of the work to date in this area has been dedicated to the habituation in the context of social robotics and human–computer interaction to achieve adaptive and realistic behavior. For instance, the ASMO architecture enables the robot to ignore irrelevant (but salient) stimuli. During the tracking task, the robot may be easily distracted by fast motion in the background. Habituation learning (implemented as a boost value attached to motion module) allows it to focus on the stimuli relevant for the task \cite{Novianto2013}. In Kismet \cite{Breazeal1999} and iCub \cite{Ruesch2008} habituation (realized as a dedicated habituation map) causes the robot to look at non-preferred or novel stimuli. In LIDA and Pogamut habituation is employed to reduce the emotional response to repeated stimuli in human-robot interaction \cite{Franklin2000d} and virtual agent \cite{Gemrot2009} scenarios respectively. MusiCog includes habituation effects to introduce novelty in music generation by aggressively suppressing the salience of elements in working memory that have been stored for a long time \cite{Maxwell2014a}.

In comparison, little attention has been paid to sensitization. In ASMO, in addition to habituation, sensitization learning allows the social robot to focus on the motion near the tracked object even though it may be slow and not salient \cite{Novianto2013}. The SASE architecture also describes both mechanisms of non-associative learning \cite{Huang2007}. 

\subsection{Priming}
\label{section_priming}
Priming occurs when prior exposure to stimulus affects its subsequent identification and classification. Numerous experiments demonstrated the existence of priming effects with respect to various stimuli, perceptual, semantic, auditory, etc., as well as behavioral priming \cite{Higgins2014}. In Section \ref{section_4_vision} we mentioned some examples of priming in vision - spatial (STAR \cite{Kotseruba2016}) and feature priming (Kismet \cite{Breazeal2001b}, ART \cite{Grossberg2003}), which allow for more efficient processing of stimuli by biasing the visual system. 

Priming in cognitive architectures has been investigated both within applications and theoretically. For example, in HTM priming is used in the context of analyzing streams of data such as spoken language. Priming is essentially a prediction of what is more likely to happen next and may resolve ambiguities based on those expectations \cite{Hawkins2006}. Experiments with CoSy show that priming speech recognition results in a statistically significant improvement compared to a baseline system \cite{Lison2008}. Other instances of priming were shown in improving perceptual categorization (ARS/SiMA \cite{Schaat2013a}), skill transfer (SASE \cite{Zhang2007}), problem solving (DUAL \cite{Kokinov1990}) and question answering (Shruti \cite{Shastri1998}).

Several models of priming validated by human data were developed in the lexical domain and problem solving. For instance, the CLARION model of positive and negative priming in lexical decision tasks models the fact that human participants are faster at identifying sequences of related concepts (e.g. the word "butter" preceded by the word "bread") \cite{Helie2014}. Leabra captures effects of priming for the production of the English past-tense inflection \cite{Reilly2000}. Finally, Darwinian Neurodynamics confirms priming in problem solving by showing that people who were primed for a more efficient solution perform better in solving puzzles \cite{Fedor2017}.

Priming is well studied in psychology and neuroscience \cite{Wentura2014} and two major theoretical frameworks for modeling priming are found in cognitive architectures: spreading activation (ACT-R \cite{Thomson}, Recommendation Architecture \cite{Coward2011}, Shruti \cite{Shastri1998}, CELTS \cite{Faghihi2011a}, LIDA \cite{Franklin2000}, ARS/SiMA \cite{Schaat2013a}, DUAL \cite{Petkov2006a}, NARS \cite{Wang2006}) and attractor networks (CLARION \cite{Helie2014}, Darwinian Neurodynamics \cite{Fedor2017}.). The current consensus in the literature is that spreading activation has greater explanation power, but attractor networks are considered more biologically plausible \cite{Lerner2012}. In this case, it seems that the choice of the particular paradigm may also be influenced by the representation. For instance, spreading activation is found in the localist architectures, where units correspond to concepts. When a concept is invoked, a corresponding unit is activated, the activation is spread to adjacent related units, which facilitates their further use. Alternatively, in attractor networks, a concept is represented by a pattern involving multiple units. Depending on the correlation (relatedness) between the patterns, activating one leads to increase in activation of others. An alternative explanation of priming is offered by SPA based on parallel constraint satisfaction \cite{Schroder2014}.

We also identified 19 architectures (mostly symbolic and hybrid) that do not implement any learning. In some areas of research, learning is not even necessary, for example, in human performance modeling, where accurate replication of human performance data is required instead (e.g. APEX, EPIC, IMPRINT, MAMID, MIDAS, etc.). Some of the newer architectures are still in the early development stage and may add learning in the future (e.g. ARCADIA, SiMA, Sigma). 

\section{Reasoning} \label{sec_reasoning}

Reasoning, originally a major research topic in philosophy and epistemology, in the past decades has become one of the focal points in psychology and cognitive sciences as well. As an ability to logically and systematically process knowledge, reasoning can affect or structure virtually any form of human activity. As a result, aside from the classic triad of logical inference (deduction, induction and abduction), other kinds of reasoning are now being considered, such as heuristic, defeasible, analogical, narrative, moral, etc. 

Predictably, all cognitive architectures are concerned with practical reasoning, whose end goal is to find the next best action and perform it, as opposed to the theoretical reasoning that aims at establishing or evaluating beliefs. There is also a third option, exemplified by the Subsumption architecture, which views reasoning about actions as an unnecessary step \cite{Brooks1987} and instead pursues physically grounded action \cite{Brooks1990}. Granted, one could still argue that a significant amount of reasoning and planning is required from a designer to construct a grounded system with non-trivial capabilities. Otherwise, in the context of cognitive architectures, reasoning is primarily mentioned with regards to planning, decision-making and learning, as well as perception, language understanding and problem-solving.

One of the main challenges a human, and consequently any human-level intelligence, faces regularly is acting based on insufficient knowledge  or "making rational decisions against a background of pervasive ignorance" \cite{Pollock2007}. In addition to sparse domain knowledge, there are limitations in terms of available internal resources, such as information processing capacity. Furthermore, external constraints, such as real-time execution, may also be introduced by the task. It should be noted that even in the absence of these restrictions, reasoning is by no means trivial. Consider, for instance, the complex machinery used by Copycat and Metacat to model analogical reasoning in a micro-domain of character strings \cite{Hofstadter1994}.

So far, the most consolidated effort has been spent on overcoming the insufficient knowledge and/or resources problem in continuously changing environments. In fact, this is the goal of general-purpose reasoning systems such as the Procedural Reasoning System (PRS), the Non-Axiomatic Reasoning System (NARS), OSCAR and the Rational Agent with Limited-Performance Hardware (RALPH). PRS is one of the earliest instances of the BDI (belief-desire-intention) model. In each reasoning cycle, it selects a plan matching the current beliefs, adds it to the intention stack and executes it. If new goals are generated during the execution, new intentions are created \cite{Georgeff1989}. NARS approaches the issue of insufficient knowledge and resources by iteratively reevaluating existing evidence and adjusting its solutions accordingly. This is possible with Non-Axiomatic Logic, which associates truth values with each statement thus allowing agents to express their confidence in the belief, a feature not available in PRS \cite{Hammer2016}. RALPH tackles complex goal-driven behavior in complex domain using decision-theoretic approach. Thus, computation is synonymous with action and action with the highest utility value (based on its expected effect) is always selected \cite{Russel1988}. The OSCAR architecture explores defeasible reasoning, i.e. reasoning which is rationally compelling but not deductively valid \cite{Koons2017}. This is a more accurate representation of everyday reasoning, where knowledge is sparse, tasks are complex and there are no certain criteria for measuring success.

All the architectures mentioned above aim at creating rational agents with human-level intelligence but they do not necessarily try to model human reasoning processes. This is the goal of architectures such as ACT-R, Soar, DUAL and CLARION. In particular, Human Reasoning Module implemented in ACT-R works under the assumption that human reasoning is probabilistic and inductive (although deductive reasoning may still occur in the context of some tasks). This is demonstrated by combining a deterministic rule-based inference mechanism with long-term declarative memory with properties similar to human memory, i.e. incomplete and inconsistent knowledge. Together, the uncertainty inherent in the knowledge and production rules enable human-like reasoning behavior, which has been confirmed experimentally \cite{Nyamsuren2014}. Similarly, symbolic and sub-symbolic mechanisms in CLARION \cite{Helie2014} and DUAL \cite{Kokinov1994a} architectures are used to model and explain various psychological phenomena associated with deductive, inductive, analogical and heuristic reasoning.

As discussed in Section \ref{section_3_taxonomies}, the question of whether higher human cognition is inherently symbolic or not is still unresolved. The symbolic and hybrid architectures considered so far, view reasoning mainly as a symbolic manipulation. Given that in emergent architectures the information is represented by the weights between individual units, what kind of reasoning, if any, can they support? It appears, based on the cognitive architecture literature, that many of the emergent architectures, e.g. ART, HTM, DAC, BBD, BECCA, simply do not address reasoning, although they certainly exhibit complex intelligent behavior. On the other hand, since it is possible to establish a correspondence between the neural networks and logical reasoning systems \cite{Martins2001}, then it should also be possible to simulate symbolic reasoning via neural mechanisms. Indeed, several architectures demonstrate symbolic reasoning and planning (SASE, SHRUTI, MicroPsi). To date, one of the most successful implementations of symbolic reasoning (and other low- and high-level cognitive phenomena) in a neural architecture is represented by SPA \cite{Rasmussen2013}. Architectures like this also raise interesting questions whether it makes sense to try and segregate symbolic parts of cognition from sub-symbolic. As most existing cognitive architectures represent a continuum from purely symbolic to connectionist, the same may be true for the human cognition as well.

\section{Metacognition}
Metacognition \cite{Flavell1979}, intuitively defined as "thinking about thinking", is a set of abilities that introspectively monitor internal processes and reason about them\footnote{Due to scope limitations we did not examine other concepts related to metacognition. Discussions of the relationship between self-awareness and metacognition, the notion of 'self', consciousness and the appropriateness of using these psychological terms in computational models can be found in \cite{Cox2005,Cox2007}. }. There has been a growing interest in developing metacognition for artificial agents, both due to its essential role in human experience and practical necessity for identifying, explaining and correcting erroneous decisions. Approximately pne-third of the surveyed architectures, mainly symbolic or hybrid ones with a significant symbolic component, support metacognition with respect to decision-making and learning. Here we will focus on three most common metacognitive mechanisms, namely self-observation, self-analysis and self-regulation.

Self-observation allows gathering data pertaining to the internal operation and status of the system. The data commonly includes the availability/requirements of internal resources (AIS \cite{Hayes-Roth1995a}, COGNET \cite{Zachary2000a}, Soar \cite{Laird2012c}) and current task knowledge with associated confidence values (CLARION \cite{Sun2016a}, CoJACK \cite{Evertsz2007}, Companions \cite{Friedman2011b}, Soar \cite{Laird2012c}). In addition, some architectures support a temporal representation (trace) of the current and/or past solutions (Companions \cite{Friedman2011b}, Metacat \cite{Jensen1998}, MIDCA \cite{Dannenhauer2014}).

Overall, the amount and granularity of the collected data depend on the further analysis and application. In decision-making applications, after establishing the amount of available resources, conflicting requests for resources, discrepancies in the conditions, etc., the system may change the priorities of different tasks/plans (AIS \cite{Hayes-Roth1995a},  CLARION \cite{Sun2016a}, COGNET \cite{Zachary2000a}, CoJACK \cite{Evertsz2007}, MAMID \cite{Hudlicka2009a}, PRODIGY \cite{Veloso1998}, RALPH \cite{Ogasawara1993a}). The trace of the execution and/or recorded solutions for the past problems can improve learning. For example, repeating patterns found in past solutions may be exploited to reduce computation for similar problems in the future (Metacat \cite{Marshall1995}, FORR \cite{Epstein2008}, GLAIR \cite{Shapiro2010}, Soar \cite{Laird2012c}). Traces are also useful for detecting and breaking out of repetitive behavior (Metacat \cite{Marshall1995}), as well as stopping learning if the accuracy does not improve (FORR \cite{Epstein2008}).

Although in theory many architectures support metacognition, its usefullness has only been demonstrated in limited domains. For instance, Metacat applies metacognition to analogical reasoning in a micro-domain of strings of characters (e.g. if abc $\rightarrow$ abd; mrrjjj $\rightarrow$ ?). Self-watching allows the system to remember past solutions, compare different answers and justify its decisions \cite{Marshall1995}. In PRS, metacognition enables efficient problem solving in playing games such as Othello \cite{Russell1989a}, better control of the simulated vehicle \cite{Ogasawara1993a} and real-time adaptability to continuously changing environments \cite{Georgeff1989}. Internal error monitoring, i.e. the comparison between the actual and expected perceptual inputs, enables the SAL architecture to determine the success or failure of its actions in a simple task, such as stacking blocks, and to learn better actions in the future \cite{Vinokurov2013}. The metacognitive abilities integrated into the dialog system make the CoSy architecture more robust to variable dialogue patterns and increase its autonomy \cite{Christensen2009}. The Companions architecture demonstrates the importance of self-reflection by replicating data from the psychological experiments on how people organize, represent and combine incomplete domain knowledge into explanations \cite{Friedman2011b}. Likewise, CLARION models human data for the Metcalfe (1986) task and replicates the Gentner and Collins (1981) experiment, both of which involve metacognitive monitoring \cite{Sun2006a}.

Metacognition is required for social cognition, particularly for the skill known in the psychological literature as a Theory of mind (ToM). ToM refers to being able to acknowledge and understand mental states of other people, use the judgment of their mental state to predict their behavior and inform one's own decision making. Very few architectures support this ability. For instance, most recently, Sigma demonstrated two distinct mechanisms for ToM using as an example for several single-stage simultaneous-move games, such as the well-known Prisoner's dilemma. The first mechanism is automatic as it involves probabilistic reasoning over the trellis graph and the second is a combinatorial search across the problem space \cite{Pynadath2013}. 

PolyScheme applies ToM to perspective taking in a human-robot interaction scenario. The robot and human in this scenario are together in a room with two traffic cones and multiple occluding elements. The human gives a command to move towards a cone, without specifying which one. If only one cone is visible to the human, the robot can model the scene from the human's perspective and use this information to disambiguate the command \cite{Trafton2005b}. Another example is reasoning about beliefs in a false belief task. In this scenario, two agents A and B observe a cookie placed in a jar. After B leaves, the cookie is moved to another jar. When B comes back, A is able to reason that B still believes that cookie is still in the first jar \cite{Scally2012}.

ACT-R was used to build several models of false belief and second-order false belief tasks (answering questions of the kind "Where does Ayla think Murat will look for chocolate?") which are typically used to assess whether children have a Theory of Mind \cite{Triona2001,Barslanrugnl2004}. However, of particular interest is the recent developmental model of ToM based on ACT-R, which was subsequently implemented on a mobile robot. Several scenarios were set up to show how ToM ability can improve the quality of interaction between the robot and humans. For instance, in a patrol scenario both the robot and human receive the task of patrolling the south area of the building, but as they start, the instructions are changed to head west. When the human starts walking towards south, the robot infers that she might have forgotten about the new task and reminds her of the new changes\cite{Trafton2013}.

\section{Practical applications}

The cognitive architectures reviewed in this paper are mainly used as research tools and very few are developed outside of academia. However, it is still appropriate to talk about their practical applications, since useful and non-trivial behavior in arbitrary domains is perceived as intelligent and often used to validate the underlying theories. We seek answers to the following questions: what cognitive abilities have been demonstrated by the cognitive architectures and what particular practical tasks have they been applied to? 

\begin{figure*}
  \includegraphics[width=1.00\textwidth]{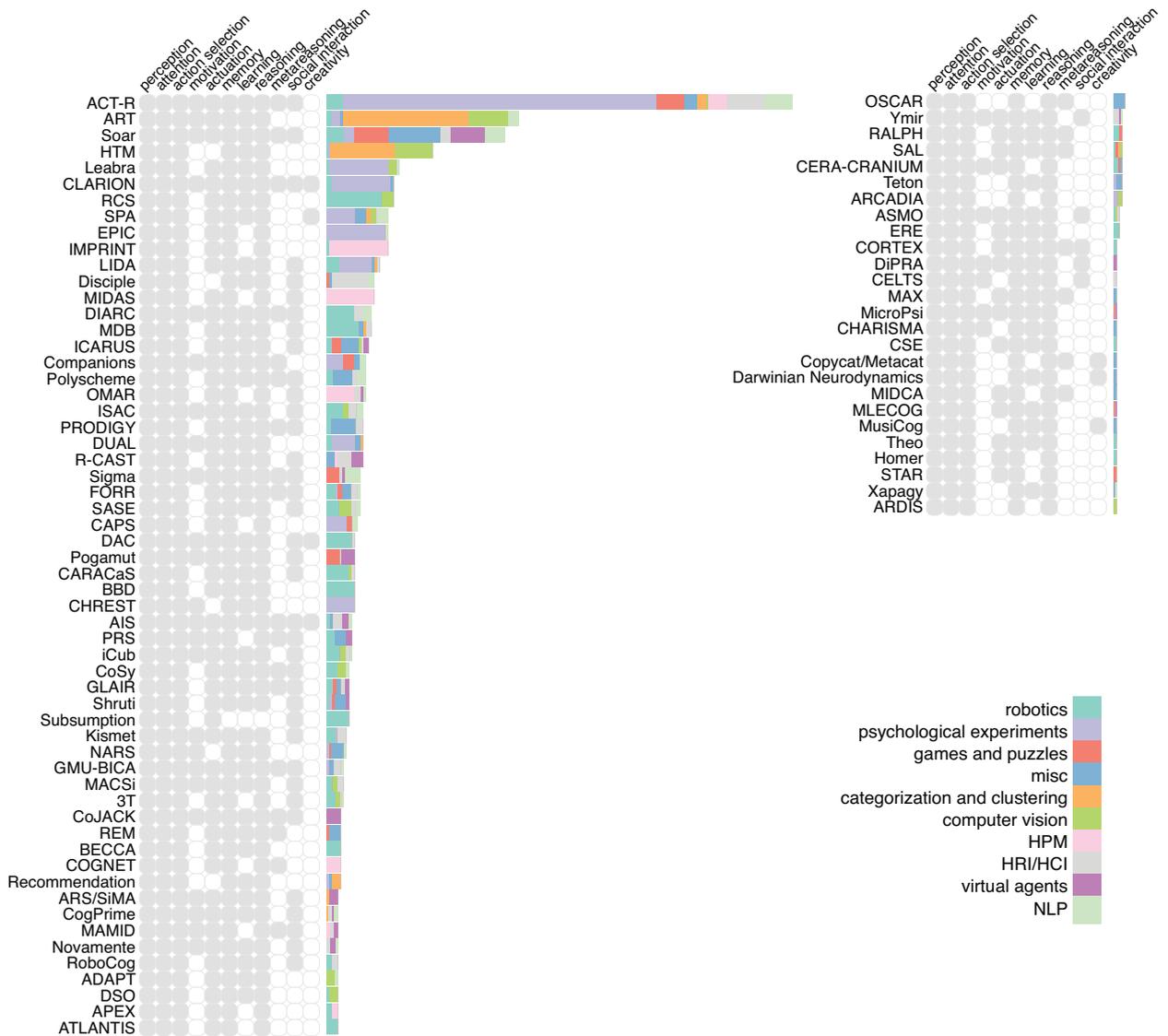}
\caption{A graphical representation of the practical applications of the cognitive architectures and corresponding competency areas defined in \cite{Adams2012a}. The plot is split into two columns for better readability (the right column being the continuation of the left one). Within each column the table shows the implemented competency areas (indicated by gray colored cells). The stacked bar plots show the practical applications: the length of the bar represents the total number of the practical applications found in the publications and colored segments within each bar represent different categories (see legend) and their relative importance (calculated as a proportion of the total number of applications implementing these categories). The architectures are sorted by the total number of their practical applications. For short descriptions of the projects, citations and statistics for each architecture refer to the interactive version of this diagram on the \href{http://jtl.lassonde.yorku.ca/project/cognitive_architectures_survey/applications.html}{project website}.}
\label{fig_10_cog_arch_applications}       
\end{figure*}

After a thorough search through the publications, we identified more than 900 projects implemented using 84 cognitive architectures. Our findings are summarized in Figure \ref{fig_10_cog_arch_applications}. It shows the cognitive abilities associated with applications, the total number of applications for each architecture (represented by the length of the bars) and what application categories they belong to (the types of categories and the corresponding number of applications in each category are shown by the color and length of the bar segments respectively). 

\subsection{Competency areas}

In order to determine the scope of the research in cognitive architectures with respect to modeling various human cognitive abilities, we began with the list of human competencies proposed by Adams et al. \cite{Adams2012a}. This list overlaps with cognitive abilities that have been identified for the purposes of evaluation of cognitive architectures (mentioned in the beginning of Section \ref{section_3_taxonomies}), and adds several new areas such as social interaction, emotion, building/creation, quantitative skills and modeling self/other.

In the following analysis, we depart from the work in \cite{Adams2012a} in several aspects. First, these competency areas are meant for evaluating AI and it is expected that each area will have a set of associated scenarios. As a result we are unable to evaluate these concepts directly, we instead will discuss which of these competency areas are represented in existing practical applications based on the published papers. Second, the authors of \cite{Adams2012a} define subareas for each competency area (e.g. tactical, strategic, physical and social planning). Although in the following sections we will review some of these subareas, they are not shown in the diagram in Figure \ref{fig_10_cog_arch_applications}. Perception, memory, attention, learning, reasoning, planning and motivation (corresponding to internal factors in Section \ref{sec_action_selection} on action selection) are covered in the previous sections and will not be repeated here. Actuation (which we simply interpret as an ability to send motor commands both in real and simulated environments)  and social interaction will be discussed in the following subsections on practical applications in robotics (Section \ref{subsection_robotics}), natural language processing (Section \ref{subsection_NLP}) and human-robot interaction (Section \ref{subsection_HRI}). Below we briefly examine the remaining competency area titled "Building/creation" (which we changed to "Creativity")\footnote{Quantitative skills, i.e. the ability to use or manipulate quantitative information, are underrepresented in cognitive architectures and therefore are excluded from our analysis. For completeness, we mention few implemented examples: counting moving and stationary objects in a simulated environment (GLAIR \cite{Santore2000}), computing a sum of two values (SPA \cite{Eliasmith2012}), identifying and counting the number of surrounding faces in a room (DIARC \cite{Scheutz2007a}) and finger counting (iCub \cite{DiNuovo2014}). Psychological experiments involving quantitative skills inlcude: counting tasks investigated by ACT-R \cite{Pape2008} and CLARION \cite{Sun2002}, cognitive arithmetic model for addition and subtraction (Soar \cite{Wang2006d}), models of multi-column subtraction used to investigate the hierarchical problem representation and solving (Teton \cite{Vanlehn1989b}, ICARUS \cite{Langley2004d} and Soar \cite{Rosenbloom1991}).}.

The \textbf{creativity} competency area in \cite{Adams2012a} involves also physical construction. Since demonstrated skills in physical construction are presently limited to toy problems, such as stacking blocks in a defined work area using a robotic arm (Soar \cite{Laird1991b}) or in simulated environments (SAL \cite{Vinokurov2013}), below we will focus on computational creativity.

Adams et al. \cite{Adams2012a} list forming new concepts and learning declarative knowledge (covered in Section \ref{section_8_learning}) within the "creation" competency area. While it can be argued that any kind of learning leads to creation of new knowledge or behavior, in this subsection we will focus more specifically on computational creativity understood as computationally generated novel and appropriate results that would be regarded as creative by humans\cite{Duch2006,Still2016}. Examples of such results are artifacts with artistic value or innovations in scientific theories, mathematical proofs, etc. Some computational models of creativity also attempt to capture and explain the psychological and neural processes underlying human creativity. 

In the domain of cognitive architectures creativity remains a relatively rare feature, for instance, only 7 out of 84 architectures on our list support creative abilities. They represent many directions in the field of computational creativity such as behavioral improvisation, music generation, creative problem solving and insight. An in-depth discussion of these problems is beyond the scope of this survey, but we will briefly discuss the contributions in these areas.

AIS and MusiCog explore improvisation in story-telling and music generation respectively. AIS enables improvisational story-telling  by controlling puppets that collaborate with humans in Virtual Theater. The core plot of the story is predefined, as well as a set of responses and emotions for each puppet. These pre-recorded responses are combined with individual perceptions, knowledge base, state information and personality of the puppets \cite{Hayes-roth1995b}. Since puppets react to both the human player and each other's actions, they demonstrate a wide range of variations on the basic plots. Such directed improvisation transforms a small set of abstract directions (e.g. be playful, curious, friendly) into novel and engaging experiences \cite{Hayes-Roth1995d}. A different strategy is followed by MusiCog.  It does not start with predefined settings, but instead listens to music samples and attempts to capture their structure. After learning is completed, the acquired structures can be used to generate novel musical themes as demonstrated on Bach BWC 1013 \cite{Maxwell2012}. Another example of synthetic music generation is Roboser, a system comprised of a mobile robot controlled by the DAC architecture and a composition engine which converts the behavioral and control states of the robot to sounds. Together, the predefined transformations and non-deterministic exploratory behavior of the robot result in novel musical structures \cite{Manzolli2005}.

Other architectures attempt to replicate the cognitive processes of human creativity rather than its results. One of the early models, Copycat (extended later to Metacat), examined psychological processes responsible for analogy-making and paradigm shift (also known as the "Aha!" phenomenon) in problem solving \cite{Marshall2002}. To uncover deeper patterns and find better analogies, Copycat performs a random micro-exploration in many directions. A global parameter controls the degree of randomness in the system and allows it to pursue both obvious and peculiar paths. Multiple simulations in a microdomain of character strings show that such approach can result in rare creative breakthroughs \cite{Hofstadter1994}.

CLARION models insight in a more cognitively plausible way and is validated on human experimental data. The sudden emergence of a potential solution is captured through interaction between the implicit (distributed) and explicit (symbolic) representations \cite{Sun2015}. For example, explicit memory search provides stereotypical semantic associations, while implicit memory allows connecting remote concepts, hence has more creative potential.  During the iterative search, multiple candidates are found through implicit search and are added into explicit knowledge. Simulations confirm that the stochasticity of implicit search improves chances of solving the problem \cite{Helie2010}.

Likewise, Darwinian Neurodynamics (DN) models insight during problem solving. In order to arrive to a solution, the system switches from analytical problem solving based on past experience to implicit generation of new hypotheses via evolutionary search. At the core of the DN model, a population of attractor networks stores the hypotheses which are selected for reproduction based on the fitness criteria. If the initial search through known answers in memory fails, the evolutionary search generates new candidates via mutations. The model of insight during solving a four-tree problem is compared to human behavioral data, accurately demonstrating the effects of priming and prior experience and the size of working memory on problem solving abilities \cite{Fedor2017}.

Last but not least is the spiking neuron model of Remote Associates Test (RAT) often used in creativity research. However, the focus is not on creativity per se, but rather on a biologically realistic representation for the words in the RAT test and associations between them using the Semantic Pointer Architecture (SPA) \cite{Gosmann2017a}.

\subsection{Application categories} \label{section_10_2_application_categories}

We identified ten major categories of applications, namely human performance modeling (HPM), games and puzzles, robotics, psychological experiments, natural language processing (NLP), human-robot and human-computer interaction (HRI/HCI), computer vision, categorization and clustering, virtual agents and miscellaneous, which included projects not related to any major group but not numerous enough to be separated into a group of their own. Such grouping of projects emphasizes the application aspect of each project, although the goal of the researchers may have been different. 

Note that some applications are associated with more than one category. For example, Soar has been used to play board games with a robotic arm \cite{Kirk2016}, which is relevant to both robotics and games and puzzles. Similarly, the ACT-R model of Tower of Hanoi evaluated against the human fMRI data \cite{Anderson2005c} belongs to games and puzzles and to psychological experiments.

\subsubsection{Psychological Experiments}
The psychological experiments category is the largest, comprising of more than one third of all applications. These include replications of numerous psychophysiological, fMRI and EEG experiments in order to demonstrate that cognitive architectures can adequately model human data or give reasonable explanations for existing psychological phenomena. If the data produced by the simulation matches the human data in some or most aspects, it is taken as an indication that a given cognitive architecture can to some extent imitate human cognitive processes. 

Most of the experiments in our list investigate psychological phenomena related to memory, perception, attention and decision making. However, there is very little repetition among the specific studies that were replicated by the cognitive architectures. Notably, Tower of Hanoi (and its derivative, Tower of London) is the only task reproduced by a handful of architectures (although not on the same human data). The task itself requires movement of a stack of disks from one rod to another and is frequently used in psychology to study problem solving and skill learning. Similarly, in various cognitive architectures ToH/ToL tasks are used to evaluate strategy acquisition (Soar \cite{Ruiz1989}, Teton \cite{Vanlehn1989e}, CAPS \cite{Just2007}, CLARION \cite{Sun2002a}, ICARUS \cite{Langley2005b}, ACT-R \cite{Anderson2001}).

Given the importance of memory for even the most basic tasks, many phenomena related to different types of memory were also examined. To name a few, the following experiments were conducted: test of the effect of working memory capacity on the syntax parsing (CAPS \cite{Just1992}), N-back task testing spatial working memory (ACT-R \cite{Kottlors2006}), reproduction of the Morris water maze test on a mobile robot to study the formation of episodic and spatial memory (BBD \cite{Krichmar2005}), effect of priming on the speed of memory retrieval (CHREST \cite{Lane2003}), effect of anxiety on recall (DUAL \cite{Feldman2009a}), etc. 

Attention has been explored both relative to perception and for general resource allocation. For example, models have been built to explain the well-known phenomena of inattentional blindness (ARCADIA \cite{Bridewell1998}) and attentional blink (LIDA \cite{Madl2012}). In addition, various dual task experiments were replicated to study and model the sources of attention limitation which reduces human ability to perform multiple tasks simultaneously. For example, CAPS proposes a functional account of attention allocation based on a dual-task experiment involving simultaneous sentence comprehension and mental rotation cite{Just2001}, ACT-R models effects of sleep loss on sustained attention performance, which requires tracking a known location on the monitor and a reaction task \cite{Gunzelmann2009}. Similar experiments have been repeated using EPIC \cite{Kieras1998,Kieras2012a}.

Multiple experiments relating perception, attention, memory and learning have been simulated using the CHREST architecture in the context of playing chess. These include investigations of gaze patterns of novice and expert players \cite{Lane2009}, effects of ageing on chess playing related to the reduced capacity of working memory and decreased perceptual abilities \cite{Smith2007} and effects of expertise, presentation time and working memory capacity on the ability to memorize random chess positions \cite{Gobet1998}.

\subsubsection{Robotics}
\label{subsection_robotics}
Nearly a quarter of all applications of cognitive architectures are related to robotics. Much effort has been spent on navigation and obstacle avoidance, which are useful on their own and are necessary for more complex behaviors. In particular, navigation in unstructured environments was implemented on an autonomous vehicle (RCS \cite{Coombs2000}, ATLANTIS \cite{Gat1992}), mobile robot (Subsumption \cite{Brooks}, CoSy \cite{Pacchierotti2006}) and unmanned marine vehicle (CARACaS \cite{Huntsberger2011b}). 

The fetch and carry tasks used to be very popular in the early days of robotics research as an effective demonstration of robot abilities. Some well-known examples include a trash collecting mobile robot (3T \cite{Firby1995a}) and a soda can collecting robot (Subsumption \cite{Brooks1989a}). Through a combination of simple vision techniques, such as edge detection and template matching, and sensors for navigation, these robots were able to find the objects of interest in unknown environments. 

More recent cognitive architectures solve search and object manipulation tasks separately. Typically, experiments involving visual search are done in very controlled environments and preference is given to objects with bright colors or recognizable shapes to minimize visual processing, for example, a red ball (SASE \cite{Weng2002a}) or a soda can (ISAC \cite{Kawamura1993}). Sometimes markers, such as printed barcodes attached to the object, are used to simplify recognition (Soar \cite{Mininger2016}). It is important to note that visual search in these cases is usually a part of a more involved task, such as learning by instruction. When visual search and localization are the end goal, the environments are more realistic (e.g. the robot controlled by CoSy finds a book on a cluttered shelf using a combination of sensors and SIFT features \cite{Lopez2008}).

Object manipulation involves arm control to reach and grasp an object. While reaching is a relatively easy problem and many architectures implement some form of arm control, gripping is more challenging even in a simulated environment. The complexity of grasping depends on many factors including the type of gripper and the properties of the object. One workaround is to experiment with grasping on soft objects, such as plush toys (ISAC \cite{Kawamura2004}). More recent work involves objects with different grasping types (objects with handles located on the top or on a side) demonstrated on a robot controlled by DIARC \cite{Wilson2016}. Another example is iCub adapting its grasp to cans of different sizes, boxes and a ruler \cite{Sauser2012}.

A few architectures implement multiple skills for complex scenarios such as a robotic salesman (CORTEX \cite{Bandera}, RoboCog \cite{Romero-Garces2015}), tutoring (DAC \cite{Vouloutsi2015}), medical assessment (LIDA \cite{Tobergte2015}, RoboCog \cite{Bandera2016}), etc. Industrial applications are represented by a single architecture - RCS, which has been used for teleoperated robotic crane operation \cite{Lytle2007}, bridge construction \cite{Bostelman1999}, autonomous cleaning and deburring workstation \cite{Murphy1988}, and the automated stamp distribution center for the US Postal Service \cite{Albus1997}.

The biologically motivated architectures focus on the developmental aspect of physical skills and sensorimotor reconstruction. For example, a childlike iCub robot platform explores acquisition of skills for locomotion, grasping and manipulation \cite{Albus1994a}, robots Dav and SAIL learn vision-guided navigation (SASE \cite{Weng2002a}) and ISAC learns grasping affordances \cite{Ulutas2008}.

\subsubsection{Human Performance Modeling (HPM)}
Human performance modeling is an area of research concerned with building quantitative models of human performance in a specific task environment. The need for such models comes from engineering domains where the space of design possibilities is too large so that empirical assessment is infeasible or too costly. 

This type of modeling has been used extensively for military applications, for example, workload analysis of Apache helicopter crew \cite{Allender2000}, modeling the impact of communication tasks on the battlefield awareness \cite{Mitchell2009a}, decision making in the AAW domain \cite{Zachary1998}, etc. Common civil applications include models of air traffic control task cite{Seamster1993}), aircraft taxi errors \cite{Wickens2005}, 911 dispatch operator \cite{Hart2001}, etc.

Overall, HPM is dominated by a handful of specialized architectures, including OMAR, APEX, COGNET, MIDAS and IMPRINT. In addition, Soar was used to implement a pilot model for large-scale distributed military simulations (TacAir-Soar \cite{Laird1998a,Jones1999}).

\subsubsection{Human-Robot and Human/Computer Interaction (HRI/HCI)}
\label{subsection_HRI}
HRI is a multidisciplinary field studying various aspects of communication between people and robots. Many of these interactions are being studied in the context of social, assistive or developmental robotics. Depending on the level of autonomy demonstrated by the robot, interactions extend from direct control (teleoperation) to full autonomy of the robot enabling peer-to-peer collaboration. Although none of the systems presented in this survey are yet capable of full autonomy, they allow for some level of supervisory control ranging from single vowels signifying direction of movement for a robot (SASE \cite{Weng1999a}) to natural language instruction (Soar \cite{Laird2012a}, HOMER \cite{Vere1990}, iCub \cite{Tikhanoff2011}). It is usually assumed that a command is of particular form and uses a limited vocabulary. 

Some architectures also target non-verbal aspects of HRI, for example, natural turn-taking in a dialogue (Ymir \cite{Thorisson2010}, Kismet \cite{Breazeal2003a}), changing facial expression (Kismet \cite{Breazeal2003c}) or turning towards the caregiver (MACsi \cite{Anzalone2013}).
 
An important practical application which also involves HCI is in building decision support systems, i.e. intelligent assistants which can learn from and cooperate with experts to solve problems in complex domains. One such domain is intelligence analysis, which requires mining large amounts of textual information, proposing a hypothesis in search of evidence and reevaluating hypotheses in view of new evidence (Disciple \cite{Tecuci2013}). Other examples include time- and resource-constrained domains such as emergency response planning (Disciple \cite{Tecuci2007a}, NARS \cite{Slam2015}), medical diagnosis (OSCAR \cite{Pollock1995b}), military operations in urban areas (R-CAST \cite{Fan2010a}) and air traffic control (OMAR \cite{Deutsch1998c}). 

\subsubsection{Natural Language Processing (NLP)}
\label{subsection_NLP}
Natural language processing (NLP) is a broad multi-disciplinary area which studies understanding of written or spoken language. In the context of cognitive architectures, many aspects of NLP have been considered, from low-level auditory perception, syntactic parsing and semantics to conversation in limited domains. 

As has been noted in the section on perception, there are only a few models of low-level auditory perception. For instance, models based on Adaptive Resonance Theory (ART), namely ARTPHONE \cite{Grossberg2003}, ARTSTREAM \cite{Grossberg1999b}, ARTWORD \cite{Grossberg1999} and others, have been used to model perceptual processes involved in speech categorization, auditory streaming, source segregation (also known as the "cocktail party problem") and phonemic integration.

Otherwise, most research in NLP is related to investigating aspects of the syntactic and semantic processing of textual data. Some examples include anaphora resolution (Polyscheme \cite{Kurup2011}, NARS \cite{Kilic2015}, DIARC \cite{Williams2016}), learning English passive voice (NARS \cite{Kilic2015}), models of syntactic and semantic parsing (SPA \cite{Stewart2015}, CAPS \cite{Just1992}) and word sense disambiguation (SemSoar and WordNet \cite{Jones2016}).

One of the early models for real-time natural language processing, NL-Soar, combined syntactic knowledge and semantics of simple instructions for the immediate reasoning and tasks in blocks world \cite{Lewis1992}. A more recent model (currently used in Soar) is capable of understanding commands, questions, syntax and semantics in English and Spanish \cite{Lindes2016}. A complete model of human reading which simulated gaze patterns, sequential and chronometric characteristics of human readers, semantic and syntactic analysis of sentences, recall and forgetting was built using the CAPS architecture \cite{Thibadeau1982}.

More commonly, off-the-shelf software is used for speech recognition and parsing, which helps to achieve a high degree of complexity and realism. For instance, a salesman robot (CORTEX \cite{Bustos2016}) can understand and answer questions about itself using a Microsoft Kinect Speech SDK. The Playmate system based on CoSy uses dedicated software for speech processing \cite{Lison2008} and can have a meaningful conversation in a subset of English about colors and shapes of objects on the table. The FORR architecture uses speech recognition for the task of ordering books from the public library by phone. The architecture, in this case, increases the robustness of the automated speech recognition system based on an Olympus/RavenClaw pipeline (CMU) \cite{Epstein2012c}.

In general, most existing NLP systems are limited both in their domain of application and in terms of the syntactic structures they can understand. Recent architectures, such as DIARC, aim at supporting more naturally sounding requests like “Can you bring me something to cut a tomato?”, however, they are still in the early stages of development \cite{Sarathy2016}. 

\subsubsection{Categorization and clustering}

Categorization, classification, pattern recognition and clustering are common ways of extracting general information from large datasets. In the context of cognitive architectures, these methods are useful for processing noisy sensory data. Applications in this group are almost entirely implemented by the emergent architectures, such as ART and HTM, which are used as sophisticated neural networks. The ART networks, in particular, have been applied to classification problems in a wide range of domains: movie recommendations (Netflix dataset \cite{Carpenter2010}), medical diagnosis (Pima-Indian diabetes dataset \cite{Kaylani2010}), fault diagnostics (pneumatic system analysis \cite{Demetgul2009}), vowel recognition (Peterson and Barney dataset \cite{Ames2008}), odor identification \cite{Distante2000}, etc. The HTM architecture is geared more towards the analysis of time series data, such as predicting IT failures\footnote{\url{grokstream.com}}, monitoring stocks\footnote{\url{numenta.com/htm-for-stocks}}, predicting taxi passenger demand \cite{Cui2015} and recognition of cell phone usage type (email, call, etc.) based on the pressed key pattern \cite{Melis2009}.
 
A few other examples from the non-emergent architectures include gesture recognition from tracking suits (Ymir \cite{Cassell1999}), diagnosis of the failures in a telecommunications network (PRS \cite{Rao1991}) and document categorization based on the information about authors and citations (CogPrime \cite{Harrigan2014}).

\subsubsection{Computer vision}

The emergent cognitive architectures are also widely applied to solving typical computer vision problems. However, these are mainly standalone examples, such as hand-written character recognition (HTM \cite{Thornton2008,Stolc2010a}), image classification benchmarks (HTM \cite{Zhuo2012,Mai2013}), view-invariant letter recognition (ART \cite{Fazl2009}), texture classification benchmarks (ART \cite{Wang1997}), invariant object recognition (Leabra \cite{OReilly2014}), etc. 

The computer vision applications that are part of more involved tasks, such as navigation in robotics, are discussed in the relevant sections.

\subsubsection{Games and puzzles}
Playing games has been an active area of research in cognitive architectures for decades. Some of the earliest models for playing tic-tac-toe and Eight Puzzle as a demo of reasoning and learning abilities were created in the 1980s (Soar \cite{Laird1984}). The goal is typically not   to master the game but rather to use it as a step towards solving similar but more complex problems. For example, Liar's Dice, a multi-player game of chance, is used to assess the feasibility of reinforcement learning in large domains (Soar \cite{Derbinsky2012c}). Similarly, playing Backgammon was used to model cognitively plausible learning in (ACT-R \cite{Sanner2000}) and tic-tac-toe to demonstrate ability to learn from instruction (Companions \cite{Hinrichs2013}). Multiple two-player board games with conceptual overlap like tic-tac-toe, the Eight Puzzle and the Five Puzzle can also be used as an effective demonstration of knowledge transfer (e.g. Soar \cite{Kirk2016}, FORR \cite{Epstein2001}). 

Compared to the classic board games used in AI research since its inception, video games provide a much more varied and challenging domain. Like board games, most video games require certain cognitive skills from a player, thus allowing the researchers to break down the problem of solving general intelligence into smaller chunks and work on them separately. However, the graphics, complexity and response times of the recent video games are getting better and better, hence many games can already be used as sensible approximations of the real world environments. In addition to that, the simulated intelligent entity is much cheaper to develop and less prone to damage than the one embodied in a physical robot. A combination of these factors makes video games a very valuable test platform for models of human cognition. 

The only drawback of using video games for research is that embedding a cognitive architecture within it requires software engineering work. Naturally, games with open source engines and readily available middleware are preferred. One such example is the Unreal Tournament 2004 (UT2004) game, for which the  Pogamut architecture \cite{Kadlec2009a} serves as a middleware, making it easier to create intelligent virtual characters. Although Pogamut itself implements many cognitive functions, it is also used with modifications by other groups to implement artificial entities for UT2004 \cite{Mora2015,Small2009,Cuadrado2009,Wang2009d,VanHoorn2009}. Other video games used in cognitive architectures research are Freeciv (REM \cite{Ulam2004}), Atari Frogger II (Soar \cite{Wintermute2012}), Infinite Mario (Soar \cite{Mohan2009}), browser games (STAR \cite{Kotseruba2016}) and custom made games (Soar \cite{Marinier2009}). It should be noted that aside from the playing efficiency and achieved scores the intelligent agents are also evaluated based on their believability (e.g. 2K BotPrize Contest\footnote{\url{http://botprize.org}}).

\subsubsection{Virtual agents}
\label{subsection_virtual_agents}
Although related to human performance modeling, the category of virtual agents is broader. While HPM requires models which can closely model human behavior in precisely defined conditions, the goals of creating virtual agents are more varied. Simulations and virtual reality are frequently used as an alternative to the physical embodiment. For instance, in the military domain, simulations model behavior of soldiers in dangerous situations without risking their lives. Some examples include modeling agents in a suicide bomber scenario (CoJACK \cite{Ritter2009}), peacekeeping mission training (MAMID \cite{Hudlicka2002a}), command and control in complex and urban terrain (R-CAST \cite{Fan2006}) and tank battle simulation (CoJACK \cite{Ritter2012}).

Simulations are also common for modeling behaviors of intelligent agents in civil applications. One of the advantages of virtual environments is that they can provide information about the state of the agent at any point in time. This is useful for studying the effect of emotions on actions, for example, in the social interaction context (ARS/SiMA \cite{Schaat2013}), or in learning scenarios, such as playing fetch with a virtual dog (Novamente \cite{Heljakka2007}).

Intelligent characters with engaging personalities can also enhance user experience and increase their level of engagement in video games and virtual reality applications. One of the examples is virtual reality drama "Human Trials" which lets human actors participate in an immersive performance together with multiple synthetic characters controlled by the GLAIR architecture \cite{Shapiro2005a,Anstey2007}). A similar project called Virtual Theater allowed users to interact with virtual characters on the screen to improvise new stories in real time. The story unfolds as users periodically select among directional options appearing on the screen (AIS \cite{Hayes-Roth1995d}).

\section*{Discussion}

The main contribution of this survey is in gathering and summarizing information on a large number of cognitive architectures influenced by various disciplines (computer science, cognitive psychology, philosophy and neuroscience). In particular, we discuss common approaches to modeling important elements of human cognition, such as perception, attention, action selection, learning, memory and reasoning. In our analysis, we also point out what approaches are more successful in modeling human cognitive processes or exhibiting useful behavior. In order to evaluate practical aspects of cognitive architectures we categorize their existing practical applications into several broad categories. Furthermore, we map these practical applications to a number of competency areas required for human-level intelligence to assess the current progress in that direction. This map may also serve as an approximate indication of what major areas of human cognition have received more attention than others. Thus, there are two main outcomes of this review. First, we present a broad and inclusive snapshot of the progress made in cognitive architectures research over the past four decades. Second, by documenting the wide variety of tested mechanisms that may help in developing explanations and models for observed human behavior, we inform future research in cognitive science and in the component disciplines that feed it. In the remainder of this section we will summarize the main themes discussed in the present survey and pertaining issues which can guide directions for future research.

\textbf{Approaches to modeling human cognition.} Historically, psychology and computer science were inspirations for the first cognitive architectures, i.e. the theoretical models of human cognitive processes and corresponding software artifacts that allowed demonstrating and evaluating the underlying theory of mind. Agent architectures, on the other hand, focus on reproducing a desired behavior and particular application demands without the constraints of cognitive plausibility. 

While often described as opposing, in practice the boundary between the two paradigms is not well defined (which causes inconsistencies in the survey literature as discussed in Section \ref{section_what_are_CA}). Despite the differences in the theory and terminology both cognitive and agent architectures have functional modules corresponding to human cognitive abilities and tackle the issues of action selection, adaptive behavior, efficient data processing and storage. For example, methods of action selection widely used in robotics and classic AI, ranging from priority queue to reinforcement learning \cite{Pirjanian1999}, are also found in many cognitive architectures. Likewise, some agent architectures incorporate cognitively plausible structures and mechanisms.

Biologically and neurally plausible models of human mind  aim to explain how known cognitive and behavioral phenomena arise from low-level brain processes. It has been argued that their support for inference and general reasoning is inadequate for modeling all aspects of human cognition, although some of the latest neuronal models demonstrate that it is far from being proven. 

The newest paradigm in AI is represented by machine learning methods, particularly data-driven deep learning, which have found enormous practical success in limited domains and even claim some biological plausibility. The bulk of work in this area is on perception although there have been attempts at implementing more general inference and memory mechanisms. The results of these efforts are stand-alone models that have yet to be incorporated within a unified framework. At present these new techniques are not widely incorporated into existing cognitive architectures. 

\textbf{Range of supported cognitive abilities and remaining gaps.}
The list of cognitive abilities and phenomena covered in this survey is by no means exhaustive. However, none of the systems we reviewed is close to supporting in theory or demonstrating in practice even this restricted subset, let alone a set of identified cognitive abilities (a comprehensive survey by Carroll \cite{Carroll1993} lists nearly 3000 of them). 

Most effort thus far has been dedicated to studying high-level abilities such as action selection (reactive, deliberative and, to some extent, emotional), memory (both short and long-term), learning (declarative and non-declarative) and reasoning (logical inference, probabilistic and meta-reasoning).  Decades of work in these areas, both in traditional AI research and cognitive architectures, resulted in creation of multiple algorithms and data structures.  Furthermore, various combinations of these approaches have been theoretically justified and practically validated in different architectures, albeit under restricted conditions. 

We only briefly cover abilities that rely on the core set of perception, action-selection, memory and learning. At present, only a fraction of the architectures have the theoretical and software/hardware foundation to support creative problem solving, communication and social interaction (particularly in the fields of social robotics and HCI/HRI), natural language understanding and complex motor control (e.g. grasping). 

With respect to the range of abilities and realism of the scenarios there remains a significant gap between the state-of-the-art in specialized areas of AI and the research in these areas within the cognitive architectures domain. Arguably some of the AI algorithms demonstrate achievements that are already on par with or even exceed human abilities (e.g. in image classification \cite{He2014}, face recognition \cite{Taigman2014}, playing simple video games \cite{Mnih2015}), which raises the bar for cognitive architectures.

Due to the complexity of modeling the human mind, groups of researchers have been pursuing complementary lines of investigation. Owing to tradition, work on high-level cognitive architectures is still centered mainly on reasoning and planning in simulated environments, downplaying issues of cognitive control of perception, symbol grounding and realistic attention mechanisms. On the other hand, embodied architectures must deal with the real-world constraints and thus pay more attention to low-level perception-motor coupling, while most of the deliberative planning and reasoning effort is spent on navigation and motor control. There are, however, trends towards overcoming these differences as classical cognitive architectures experiment with embodiment \cite{Mohan2012,Trafton2013} or are being interfaced with robotic architectures \cite{Scheutz2013a} to utilize their respective strengths.

There also remains a gap between the biologically inspired architectures and neural simulations and the rest of the architectures. Even though these models can demonstrate how low-level neural processes give rise to some higher-level cognitive functions, they do not have the same range and efficiency in practical applications compared to the less theoretically restricted systems. Some exceptions exist, for example Grok - a commercial application for IT analytics based on the biologically inspired HTM architecture or Neural Information Retrieval System implemented as a hierarchy of ART networks for storing 2D and 3D parts designs built for the Boeing company \cite{Smith1997}. 

Finally, most of the featured architectures cannot reuse the capabilities or accumulate knowledge as they are applied to new tasks. Instead, every new task or skill is demonstrated using a  separate model, specific set of parameters or knowledge base \footnote{Note that the proposed visualizations demonstrate aggregated statistics and do not imply that there exists a single model that encapsulates them all.}. Examples of architectures capable of switching between various scenarios/environments without resetting or changing parameters are sparse and are explicitly declared as such, e.g. the SPA architecture which can perform eight unrelated tasks \cite{Eliasmith2012}.

\textbf{Outstanding issues and future directions.}
One of the outcomes of this survey, besides summarizing past and present attempts at modeling various human cognitive abilities, is highlighting the problems needing further research via \href{http://jtl.lassonde.yorku.ca/project/cognitive_architectures_survey/applications.html}{interactive visualization}. We also reviewed the future directions suggested by scholars in a large number of publications in search of problems deemed important. We present some of the main themes below, excluding architecture- and implementation-specific issues.

\textbf{Adequate experimental validation.} 
A large number of papers end in a call for thorough experimental testing of the given architecture in more diverse, challenging and realistic environments \cite{Firby1995a,Rohrer2012,Manso2016} and real-world situations \cite{Sun2015} using more elaborate scenarios \cite{Rousseau1996a} and diverse tasks \cite{Herd2013}. This is by far the most pressing and long-standing issue (found in papers from the early 90s \cite{Matthies1992} up until 2017 \cite{Fedor2017}) which remains largely unresolved. The data presented in this survey (e.g. Section \ref{section_10_2_application_categories} on practical applications) shows that, by and large, the architectures (including those vying for AGI) have practical achievements only in several distinct areas and in highly controlled environments. Likewise, common validation methods have many limitations. For instance, replicating psychological experiments, the most frequently used in our sample of architectures, presents the following challenges: it is usually based on a small set of data points from human participants, the amount of prior knowledge for the experiment is often unknown and modeled in an ad hoc manner, pre-processing of stimuli is required in case the architecture does not posess adequate perception, and, furthermore, the task definition is narrow and tests only a particular aspect of the ability \cite{Jones2012}. Overall, such restricted evaluation tells us little about the actual abilities of the system outside the laboratory setting. 

\textbf{Realistic perception.} 
The problem of perception is most acute in robotic architectures operating in noisy unstructured environments. Frequently mentioned issues inlcude the lack of active vision \cite{Huber1995}, accurate localization and tracking \cite{Wolf2010}, robust performance under noise and uncertainty \cite{Romero-Garces2015} and the utilization of context information to improve detection and localization \cite{Christensen2009}. More recently, deep learning is being considered as a viable option for bottom-up perception \cite{Oudeyer,Bona2013}. 

These concerns are supported by our data as well. Overall, in comparison to higher-level cognitive abilities, the treatment of perception, and vision in particular, is rather superficial, both in terms of capturing the underlying processes and practical applications. For instance, almost half of the reviewed architectures do not implement any vision (Figure \ref{fig_5_cog_arch_vision}). The remaining projects, with few exceptions, either default to simulations or operate in controlled environments. 

Other sensory modalities, such as audition, proprioception and touch, often rely on off-the-shelf software solutions or are trivially implemented via physical sensors/simulations. Multi-modal perception in the architectures that support it, is approached mechanistically and is tailored for a particular application. Finally, bottom-up and top-down interactions between perception and higher-level cognition are largely overlooked both in theory and practice. As we discuss in Section \ref{section_5_attention} many of the mechanisms are a by-product of other design decisions rather than a result of deliberate efforts. 

\textbf{Human-like learning.}
Even though various types of learning are represented in the cognitive architectures, there is still a need for developing more robust and flexible learning mechanisms \cite{Hinrichs2013,Menager2016}, knowledge transfer \cite{Herd2013} and accumulation of new knowledge without affecting prior learning \cite{Jonsdottir2013}. Learning is especially important for developmental approach to modeling cognition, specifically motor skills and new declarative knowledge (as discussed in Section \ref{section_8_learning}). However, artificial development is restricted and far less efficient because visual and auditory perception in robots is not yet on par with even the small children \cite{Christensen2009}.

\textbf{Natural communication.}
Verbal communication is the most common mode of interaction between the artificial agents and humans. Many issues are yet to be resolved as current approaches do not possess a sufficiently large knowledge base for generating dialogues \cite{Nkambou2013} and generally lack robustness \cite{Hinrichs2013,Williams2015}. Few, if any, architectures are capable of detecting the emotional state and intentions of the interlocutor \cite{Breazeal2003} or give personalized responses \cite{Gobet2012}. Non-verbal aspects of communication, such as performing and detecting gestures and facial expressions, natural turn-taking, etc., are investigated by only a handful of architectures, many of which are no longer in development. As a result, demonstrated interactions are heavily scripted and often serve as voice control interfaces for supplying instructions instead of enabling collaboration between human and the machine \cite{Myers2002c} (see also discussion in Section \ref{subsection_NLP}). 

\textbf{Autobiographic memory.}
Even though, memory, a necessary component of any computational model, is well represented and well researched in the domain of cognitive architectures, episodic memory is comparatively under-examined \cite{Danker2010,Ladislau2012,Borrajo2015,Menager2016}. The existence of this memory structure has been known for decades \cite{Tulving1972} and its importance for learning, communication and self-reflection widely recognized, however episodic memory remains relatively neglected in computational models of cognition (see Section \ref{fig_8_cog_arch_memory}). This is also true for general models of human mind such as Soar and ICARUS, which included it relatively recently. 

Currently the majority of the architectures save episodes as time-stamped snapshots of the state of the system. This approach is suitable for agents with a short life-span and a limited representational detail but more elaborate solutions will be needed for life-long learning and managing large-scale memory storage \cite{Laird2009a}. These issues are relevant to other types of long-term memory as well \cite{Murdock2008,Henderson2012,Bellas2014}. 

\textbf{Computational performance.}
The problem of computational efficiency, for obvious reasons, is more pressing in robotic architectures \cite{Gat1994,Wilson2014a} and interactive applications \cite{Ash1996,Moreno2006}. However, it is also reported for non-embodied architectures \cite{Sun2016a,Jones2016}, particularly neural simulations \cite{Wendelken2004,Tripp2016}. Overall, time and space complexity is glossed over in the literature, frequently described in qualitative terms (e.g. 'real-time' without specifics) or omitted from discussion altogether. 

A related point is an issue of scale. It has been shown that scaling up existing algorithms to larger amounts of data introduces additional challenges, such as the need for more efficient data processing and storage. For example, the difficulties associated with maintaining a knowledge base equivalent to human brain memory-storage capacity are discussed in Section \ref{section_7_memory}. These problems are currently addressed by only a handful of architectures, even though solving them is crucial for further development of theoretical and applied AI.
  
In addition to concerns collectively expressed by the researches (and supported by the data in this survey) below we examine two more observations: 1) the need for an objective evaluation of the results and measuring the overall progress towards understanding and replicating human-level intelligence and 2) reproducibility of the research in the cognitive architectures. 

\textbf{Comparative evaluation of cognitive architectures.}
The resolution of the commonly acknowledged issues hinges on the development of objective and extensive evaluation procedures. We have already mentioned as one of the issues that individual architectures are validated using a disjointed and sparse set of experiments, which cannot support claims about the abilities of the system or adequately assess its representational power. However, comparisons between architectures and measuring overall progress of the field entails an additional set of challenges.

Surveys are a first step towards mapping the landscape of approaches to modeling human intelligence, but they can only identify the blank spots needing further exploration and are ill-suited for comparative analysis. For instance, in this paper, visualizations show only the existence of any particular feature, be it memory, learning method or perception modality. However, they cannot represent the extent of the ability nor whether it is enabled by the architectural mechanisms or is an ad hoc implementation\footnote{Refer to a relevant discussion in Section \ref{section_7_memory} on how working memory size is set in various architectures.}. Thus, any comparisons between the architectures are discouraged and any such conclusions should be taken with caveats. 

A compounding factor for evaluation is that cognitive architectures comprise both theoretical and engineering components, which leads to a combinatorial explosion of possible implementations. Consider, for example, the theoretical and technical nuances of various instantiations of the global workspace concept, e.g. ARCADIA \cite{Bridewell2015}, ASMO \cite{Novianto2010} and LIDA \cite{Franklin2012} or the number of planning algorithms and approaches to dynamic action selection in Section \ref{sec_action_selection}. Furthermore, building a cognitive architecture is a major commitment, requiring years of development to put even the basic components together (average age of projects in our sample is $\approx15$ years), making any major changes harder with time. Thus, evaluation is crucial for identifying more promising approaches and pruning suboptimal paths for exploration. To some extent, it has already been happening organically via a combination of theoretical considerations and collective experience gathered over the years. For instance, symbolic representation, once dominant, is being replaced by more flexible emergent approaches (see Figure \ref{fig_2_cog_arch_timeline}) or augmented with sub-symbolic elements (e.g. Soar, REM). Clearly, more consistent and directed effort is needed to make this process faster and more efficient. 

Ideally, proper evaluation should target every aspect of cognitive architectures using a combination of theoretical analysis, software testing techniques, benchmarking, subjective evaluation and challenges. All of these have already been sparingly applied to individual architectures. For example, a recent theoretical examination of the knowledge representations used in major cognitive architectures reveals their limitations and suggests ways to amend them \cite{Lieto2016,Lieto2017}. Over the years, a number of benchmarks and challenges have been developed in other areas of AI and a few have been already used to evaluate cognitive architectures, e.g. 2K BotPrize competition (Pogamut \cite{Gemrot2010}), various benchmarks, including satellite image classification (ART \cite{Carpenter2010}), medical data analysis (ART \cite{Carpenter}) and face recognition  (HTM \cite{Stolc2010a}), as well as robotics competitions such as AAAI Robot Competition (DIARC \cite{Schermerhorn2006} and RoboCup (BBD \cite{Szatmary2006}). 

Tests probing multiple abilities via series of tasks are suggested for evaluating cognitive architectures and general-purpose AI systems. Some examples include, "Cognitive Decathlon" \cite{Mueller2008} and  I-Athlon \cite{Adams2016}, both inspired by the Turing test, as well as Catell-Horn-Carroll model examining factors of intelligence \cite{Ichise2016} or a battery of tests for assessing different cognitive dimensions proposed by Samsonovich et al. \cite{Samsonovich2006}. However, most of the work in this area remains theoretical and has not resulted in concrete proposals ready to be implemented or complete frameworks.
For further discussion of ample literature concerning evaluation of AI and cognitive architectures we refer the readers to a very recent and comprehensive review of these methods in \cite{Hernandez-Orallo2017}. 

\textbf{Reproducibility of the results.}
Overall, in the literature on cognitive architectures far more importance is given to the cognitive, psychological or philosophical aspects, while technical implementation details are often incomplete or missing. For example, in Section \ref{section_4_vision} on visual processing we list architectures that only briefly mention some perceptual capabilities but do not specify enough concrete facts for analysis. Likewise, justifications of using particular algorithms/representations along with their advantages and limitations are not always disclosed. Generally speaking, lack of full technical detail compromises reproducibility of the research.

In part, these issues can be alleviated by providing access to the software/source code, but only one-thirds of the architectures do so. To a lesser extent this applies to architectures such as BBD, SASE, Kismet, Ymir, Subsumption and a few others that are physically embodied and depend on a particular robotic platform. However, the majority of architectures we reviewed have a substantial software component, releasing which could be of benefit to the community and the researchers themselves. For instance, cognitive architectures such as ART, ACT-R, Soar, HTM and Pogamut, are used by many researchers outside the main developing group.

In conclusion, our hope is that this work will serve as an overview and a guide to the vast field of cognitive architecture research as we strived to objectively and quantitatively assess the state-of-the-art in modeling human cognition.

\bibliographystyle{spmpsci}      


%
%

\end{document}